\documentclass[lettersize,journal]{IEEEtran}
\usepackage{amsmath,amsfonts}
\usepackage{algorithmic}
\usepackage{algorithm}
\usepackage{array}
\usepackage[caption=false,font=normalsize,labelfont=sf,textfont=sf]{subfig}
\usepackage{textcomp}
\usepackage{stfloats}
\usepackage{url}
\usepackage{verbatim}
\usepackage{graphicx}
\usepackage{bbm}
\usepackage{cite}
\usepackage{amsmath,amssymb,amsthm}
\usepackage{orcidlink}
\usepackage{multirow}
\usepackage{makecell}
\usepackage{booktabs} 
\hypersetup{hidelinks,
	colorlinks=true,
	allcolors=blue,
	pdfstartview=Fit,
	breaklinks=true}
\usepackage{tcolorbox}
\usepackage{xspace}
\usepackage{colortbl}
\newcommand{\method}{\texttt{\textbf{SENT}}\xspace}

\definecolor{darkred}{rgb}{0.55, 0.0, 0.0}
\definecolor{darkgreen}{rgb}{0.0, 0.55, 0.0}
\newcommand{\codesite}{\url{https://github.com/zldscr0/SENT}}

\begin{document}

\title{Efficient Reinforcement Learning with Semantic and Token Entropy for LLM Reasoning}

\author{
Hongye Cao\orcidlink{0000-0002-2537-2295}, Zhixin Bai\orcidlink{0009-0002-5877-5055}, Ziyue Peng\orcidlink{0009-0002-9935-0758}, Boyan Wang\orcidlink{0000-0003-0088-2106}, Tianpei Yang\orcidlink{0000-0002-5497-7146}, Jing Huo\orcidlink{0000-0002-8504-455X}, \\ Yuyao Zhang, and Yang Gao\orcidlink{0000-0002-2488-1813},~\IEEEmembership{Senior Member,~IEEE,}

\thanks{
Hongye Cao, Zhixin Bai, Ziyue Peng, Boyan Wang, Tianpei Yang, Jing Huo, and Yang Gao are with the National Key Laboratory for Novel Software Technology, Nanjing University, Nanjing 210093, China (e-mail: hongyecao528@gmail.com; 652024330001@smail.nju.edu.cn; 231220093@smail.nju.edu.cn; boyanwang@nju.edu.cn; tianpei.yang@nju.edu.cn; huojing@nju.edu.cn;  gaoy@nju.edu.cn). \textit{(Hongye Cao and Zhixin Bai contributed equally to this work.)}
}
\thanks{
Yuyao Zhang is with China Mobile NineVerse Artificial Intelligence Technology (Beijing) Co., Ltd., and Institute of Artificial Intelligence, NineVerse, Beijing 100032, China (e-mail: zhangyuyao@cmjt.chinamobile.com).
}
}



\maketitle

\begin{abstract}
Reinforcement learning with verifiable rewards (RLVR) has demonstrated superior performance in enhancing the reasoning capability of large language models (LLMs). However, this accuracy-oriented learning paradigm often suffers from entropy collapse, which reduces policy exploration and limits reasoning capabilities. To address this challenge, we propose an efficient reinforcement learning framework that leverages entropy signals at both the semantic and token levels to improve reasoning. From the data perspective, we introduce semantic entropy-guided curriculum learning, organizing training data from low to high semantic entropy to guide progressive optimization from easier to more challenging tasks. For the algorithmic design, we adopt non-uniform token treatment by imposing KL regularization on low-entropy tokens that critically impact policy exploration and applying stronger constraints on high-covariance portions within these tokens. By jointly optimizing data organization and algorithmic design, our method effectively mitigates entropy collapse and enhances LLM reasoning. Experimental results across $6$ benchmarks with $3$ different parameter-scale base models demonstrate that our method outperforms other entropy-based approaches in improving reasoning.
\end{abstract}


\section{Introduction}
\IEEEPARstart{R}{easoning} has emerged as a core capability of large language models (LLMs) 
in tackling complex tasks~\cite{11123142,guo2025deepseek}. Reinforcement learning with verifiable rewards (RLVR) has effectively enhanced LLMs' reasoning capabilities across mathematics~\cite{10946242,yue2025vapo}, code generation~\cite{shao2024deepseekmath,dou2024stepcoder}, and decision-making applications~\cite{10938647,shihi_robot,lin2024correctable} through post-training. However, this purely accuracy-based learning paradigm decreases exploration and may lead to local optima~\cite{yu2025dapo,cheng2025reasoning}, thereby limiting the model's ability to improve reasoning performance. This problem manifests as precipitous entropy collapse during post-training. Entropy, which quantifies uncertainty in the policy's action distribution and measures exploration capability~\cite{williams1992simple,haarnoja2018soft,eysenbach2022maximum}, drops sharply during training, resulting in generated responses that lack diversity and further limiting exploration.

To encourage exploration and improve reasoning, prior works have investigated entropy-guided strategies to enhance LLMs' reasoning. These approaches employ various techniques, including dynamic temperature coefficient adjustment based on entropy changes~\cite{liao2025enhancing}, introduction of novel reward signals for optimization~\cite{cheng2025reasoning,vanlioglu2025entropy}, suppressing tokens that contribute most to entropy decline~\cite{cui2025entropy}, and masking low-entropy tokens~\cite{wang2025beyond}.
\IEEEpubidadjcol 

\begin{figure}
    \centering
    \includegraphics[width=0.9\linewidth]{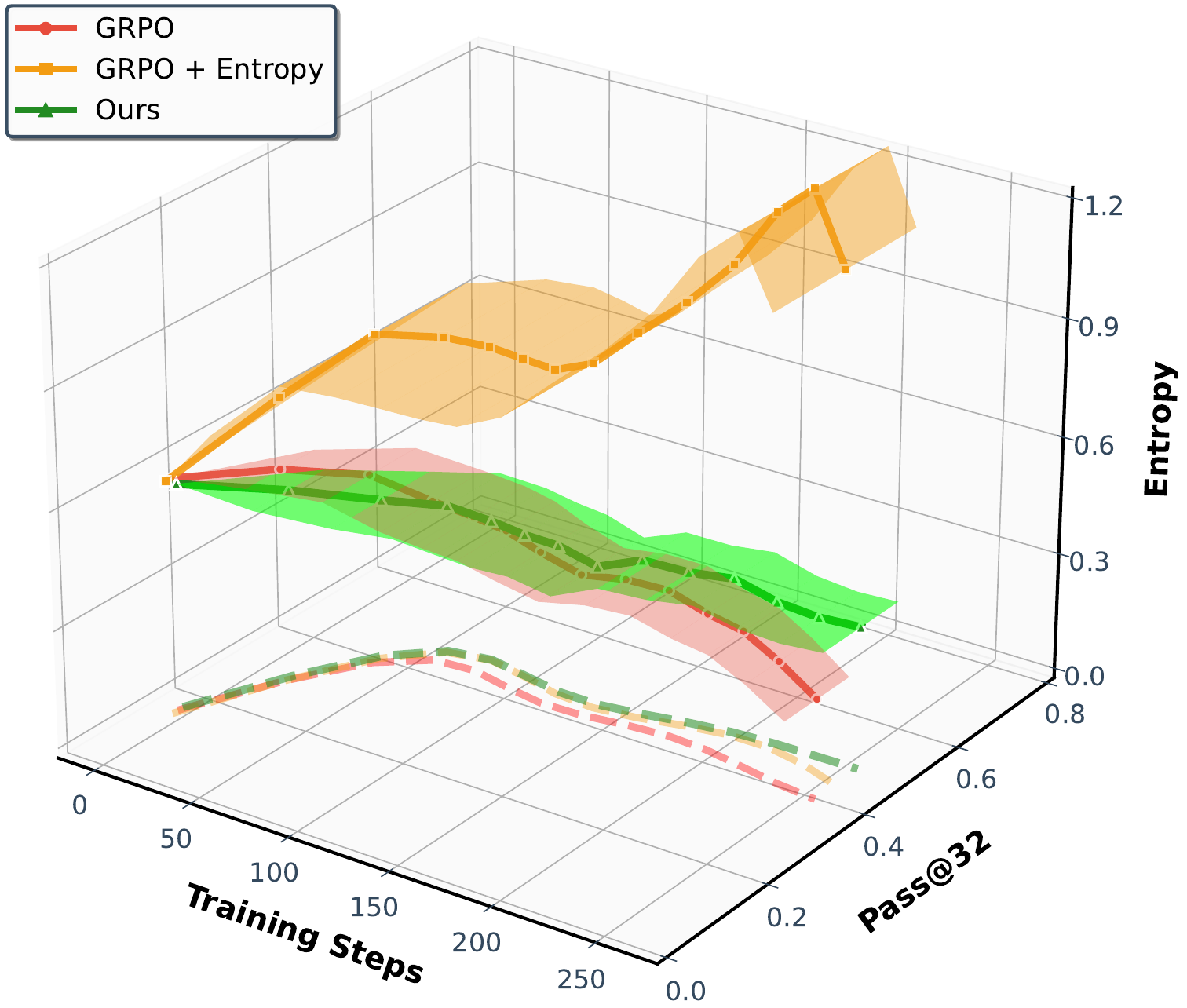}
    \caption{Learning curves in entropy and \textit{Pass@32} during training of \method compared with GRPO and GRPO with entropy. The shadow is the derivation of the performance of \textit{Pass@32} in AIME24~\cite{codeforcesamerican} benchmark.}
    \label{fig:example}
\end{figure}

While these approaches have shown promising results, they suffer from critical limitations that constrain reasoning.
As shown in Fig.~\ref{fig:example}, GRPO~\cite{shao2024deepseekmath} experiences entropy collapse that diminishes exploration capacity, whereas the method that directly incorporates an entropy-maximization objective encounters entropy explosion, resulting in policy instability. These limitations fundamentally arise from entropy fluctuations that destabilize policy learning and manifest through three key issues:
First, existing methods~\cite{cui2025entropy,cheng2025reasoning,wang2025beyond} focus on local token-level entropy characteristics within individual samples while neglecting the global perspective of training data organization. The difficulty distribution of training samples significantly influences optimization and entropy changes, yet existing works treat all samples equally regardless of their semantic complexity, leading to abrupt difficulty transitions that trigger unstable exploration performance. Second, at the algorithmic level, most approaches~\cite{vanlioglu2025entropy,liao2025enhancing} apply uniform optimization strategies across all tokens without distinguishing their varying impacts on policy exploration. This one-size-fits-all approach fails to recognize that low-entropy tokens, which directly constrain exploration, require targeted intervention distinct from high-entropy tokens. Third, current approaches lack fine-grained control within tokens. Even among low-entropy tokens, different portions exhibit varying degrees of entropy changes, but existing methods apply homogeneous constraints without adapting to these internal variations. These limitations result in suboptimal exploration strategies that not only fail to comprehensively mitigate entropy collapse but also cannot maintain stable improvement in reasoning, thereby failing to unlock the reasoning potential of LLMs.

To address these challenges, we propose an efficient reinforcement learning (RL) framework that combines data-level \textbf{S}emantic \textbf{EN}tropy with \textbf{T}oken-level entropy optimization (\method) for enhancing LLM reasoning. \method is motivated by a fundamental insight: \textbf{entropy collapse stems from both inadequate data curriculum that fails to scaffold learning complexity and token-level optimization strategies that treat all positions uniformly, ignoring the heterogeneous importance of different tokens in reasoning chains.}

At the data level, we introduce curriculum learning guided by semantic entropy (SE) that organizes training data progressively from low to high entropy, creating a learning process from simpler to complex reasoning tasks. This curriculum design prevents premature convergence by progressively increasing the difficulty of training data, allowing the model to gradually adapt to more complex reasoning tasks. For the algorithmic design, we propose that reasoning chains exhibit inherent structural heterogeneity that low-entropy tokens, which represent near-deterministic decisions, critically constrain the model's exploration capacity and contribute to entropy collapse. Rather than treating all tokens uniformly, we identify low-entropy tokens and impose KL regularization to prevent over-optimization at these parts. Importantly, we further analyze the internal structure of low-entropy tokens and apply stronger constraints specifically to high-covariance portions within these tokens, where increased uncertainty preservation yields the benefit of maintaining exploration. This fine-grained constraint mechanism encourages targeted exploration at positions that most critically affect policy diversity. Through this dual-level optimization, \method achieves sustained exploration by combining semantic entropy-based data curriculum that provides a structured difficulty progression with fine-grained token-selective KL regularization that preserves policy diversity at critical decision points, thereby enabling stable reasoning improvement during training.

The main contributions of this work can be summarized:
\begin{itemize}
\item We highlight that enhancing reasoning capabilities requires exploration-aware entropy-based strategies at both the data organization and algorithmic optimization, and we are the first to jointly optimize exploration at these two levels for LLMs' reasoning.
\item We propose \method that combines semantic entropy-based curriculum learning at the data-level with token-level low entropy optimization. \method organizes training data from low to high semantic entropy, and applies KL regularization on low-entropy tokens with stronger constraints on their high-covariance portions to encourage exploration. These two levels work in tandem so that the curriculum provides stable difficulty scaffolding while token-level constraints encourage exploration. 
\item We conduct extensive experiments across $6$ benchmarks on $1.5$B, $7$B, and $14$B base models, demonstrating that \method outperforms existing entropy-based approaches and effectively mitigates entropy collapse while enhancing LLMs' reasoning performance.
\end{itemize}

The remainder of this paper is organized as follows. Section~\ref{sec:related work} provides an overview of related works, including RL for LLMs, and entropy-based exploration in LLMs. Section~\ref{sec:preliminaries} details the necessary preliminaries. Section~\ref{sec:approach} presents the proposed approach in detail. In section \ref{sec:experiments}, we introduce the experiments conducted to demonstrate the superiority of \method. Finally, section~\ref{sec:conclusion} draws conclusions and discusses future works.

\section{Related Work}
\label{sec:related work}
\subsection{Reinforcement Learning for LLMs}
With the rapid development of LLMs, reinforcement learning with human feedback (RLHF) has been widely adopted for aligning models with human preferences~\cite{ouyang2022training,10766898}. Pioneering works such as InstructGPT~\cite{ouyang2022training} and Constitutional AI~\cite{bai2022constitutional} demonstrated the effectiveness of RLHF in improving helpfulness, harmlessness, and honesty of language models. These methods typically employ PPO~\cite{schulman2017proximal} with reward models trained on human preference data to guide policy learning.
Following the introduction of OpenAI's o1 reasoning model~\cite{o1}, RLVR has emerged as a promising paradigm for enhancing reasoning capabilities in LLMs, particularly in mathematics and programming domains~\cite{lightman2023let,shao2024deepseekmath,lambert2024tulu}. Unlike RLHF which relies on human preferences, RLVR leverages automatic verification of solutions through test cases or formal proofs, providing more reliable and scalable training signals. This breakthrough has sparked the development of numerous reasoning models, including DeepSeek R1~\cite{guo2025deepseek}, QwQ~\cite{qwq}, and Qwen3~\cite{yang2025qwen3}. DeepSeek R1 achieved substantial performance improvements through the GRPO algorithm~\cite{shao2024deepseekmath}, which addresses the computational challenges of traditional actor-critic methods by using group-based advantage estimation. Subsequently, DAPO~\cite{yu2025dapo} was proposed to address limitations in GRPO by incorporating direct advantage computation, followed by other RLVR methods such as VAPO~\cite{yue2025vapo} and GSPO~\cite{zheng2025group}, all aimed at improving LLMs' reasoning capabilities through more efficient policy optimization strategies. \textit{In this work, we adopt GRPO as our baseline to investigate the reasoning capability of LLMs.}

\subsection{Entropy-Based Exploration in LLMs}
Entropy, as a measure of uncertainty in probability distributions, has been widely recognized as a crucial signal for exploration in RL~\cite{haarnoja2018soft,eysenbach2018diversity,10737895}. As noted in DAPO~\cite{yu2025dapo}, when using simple PPO or GRPO algorithms, rapid entropy collapse is commonly observed, resulting in generated responses that lack diversity and limiting the model's exploration capacity. 
Entropy-based methods have been extensively explored in traditional decision-making tasks and have demonstrated strong performance~\cite{berrueta2024maximum,chao2024maximum,messaoud2024sac,10783450,10476495}. 
Building upon this foundation, recent works have increasingly focused on entropy-guided methods to enhance exploration in LLMs.
Dynamic resource allocation~\cite{liao2025enhancing} introduces mechanisms such as dynamic rollout budget allocation and temperature scheduler, which adaptively allocate sampling resources based on task difficulty and dynamically adjust sampling temperature to encourage exploration. Token-level covariance analysis~\cite{cui2025entropy} reveals that policy entropy changes are proportional to the covariance between action probabilities and logit changes, leading to methods that randomly prune high-covariance tokens. Beyond the 80/20 rule~\cite{wang2025beyond} identifies that high-entropy tokens play critical roles in reasoning paths and proposes policy gradient updates using only these tokens. Advantage reweighting approach~\cite{vanlioglu2025entropy} dynamically assigns advantage weights based on each token's advantage and entropy to balance exploration and exploitation. Entropy-augmented optimization~\cite{cheng2025reasoning} adds a clipped and gradient-detached entropy term to the advantage function in PPO and GRPO, encouraging longer and deeper reasoning chains under high uncertainty while preserving the original policy optimization direction.
Despite these advances, existing entropy-based methods predominantly operate at individual token level. Moreover, they focus primarily on algorithmic optimization while neglecting the role of training data organization in entropy dynamics. Our work addresses these limitations by proposing a framework that jointly optimizes entropy signals at both the data level and the token level.

\section{Preliminary}
\label{sec:preliminaries}

\subsection{Group Relative Policy Optimization (GRPO)}

GRPO is an extension of PPO to stabilize policy updates by normalizing advantage estimates over groups of samples. The key innovation is improving the robustness of policy gradients by reducing variance through group-based advantage computation, thereby eliminating the need for a separate value network. The standard PPO objective is defined as:
\begin{equation}
\begin{aligned}
    \mathcal{J}_{\mathrm{PPO} }(\theta) &   =\mathbb{E}_{(q,a)\sim \mathcal{D},o\sim \pi_{\theta_\mathrm{old}}(\cdot |q)} 
    \\ & \min \left[
r_t(\theta)A_t,\mathrm{clip}(r_t(\theta),1-\epsilon,1+\epsilon)A_t \right], 
\end{aligned}
\label{eq:ppo}
\end{equation}
where $\mathcal{D}$ is the dataset of queries $q$ and corresponding ground-truth answers $a$, $\epsilon \in \mathbb{R}$ is a hyperparameter set to $2.0$, and $A_t$ is the advantage calculated by a value network at timestep $t$. The likelihood ratio for given question $q$ and output $o$ is expressed below:
\begin{equation}
r_t(\theta)=\frac{\pi_{\theta}(o_t|q,o_{<t})}{\pi_{\theta_{\mathrm{old}}}(o_t|q,o_{<t})}  .  
\end{equation}

Building on the clipped objective in Eq.~\ref{eq:ppo}, GRPO eliminates the value network by estimating advantages using the average reward within a group of sampled responses. Specifically, for each query $q$ and its ground-truth answer $a$, the rollout policy $\pi_{\theta_{\mathrm{old}}}$ generates a group of responses $\{o^i\}^{G}_{i=1}$ with corresponding outcome rewards $\{R^i\}^{G}_{i=1}$, where $G \in \mathbb{R}$ is the group size. The estimated advantage $\hat{A}^i_t$ is computed as:
\begin{equation}    
\hat{A}^i_t=\frac{r^i-\mathrm{mean}(\{R^i\}^{G}_{i=1}) }{\mathrm{std}(\{R^i\}^{G}_{i=1}) },
\label{eq:adv}
\end{equation}
where $R^i=\begin{cases}
 1.0 & \text{ if }  \mathrm{equivalent}(a, o^i)\\
0.0  & \text{ if } \mathrm{otherwise}
\end{cases}$ .
\vspace{1.5mm}

Moreover, GRPO incorporates a KL divergence penalty to constrain policy updates and prevent the trained policy from deviating excessively from the reference policy. Hence, the GRPO objective is:
\begin{equation}
\label{eq:grpo}
    \begin{aligned}
\mathcal{J}_{\mathrm{GRPO}}(\theta) 
&= \mathbb{E}_{(q,a)\sim \mathcal{D},\,\{o^i\}_{i=1}^G \sim \pi_{\theta_{\mathrm{old}}}(\cdot | q)} 
\\
& \Bigg[ \frac{1}{G} \sum_{i=1}^{G} \frac{1}{|o^i|} 
\sum_{t=1}^{|o^i|}
\Big( \min\!\big(r^i_t(\theta)\hat{A}^i_t,\, \mathrm{clip}(r^i_t(\theta),
\\
& 1-\epsilon,\, 1+\epsilon)\hat{A}^i_t\big)
- \beta D_{\mathrm{KL}}(\pi_{\theta} \,\|\, \pi_{\mathrm{ref}}) \Big)
\Bigg],
    \end{aligned}
\end{equation}
where $\pi_{\mathrm{ref}}$ is the reference model, which is usually the initial supervised fine-tuning (SFT) model, and $\beta$ is the coefficient for the penalty.


\subsection{Entropy Calculation}
Policy entropy quantifies the uncertainty or randomness in the action distribution of a policy. Given a policy model $\pi_{\theta}$, for each token $o_t$ in an output $o$, the entropy of the current policy over the vocabulary $\mathcal{V}$ is: 
\begin{equation}
\mathcal{H}_t=-\sum_{v\in\mathcal{V} }\pi_{\theta}(v|q,o_{<t})\log \pi_{\theta}(v|q,o_{<t}). 
\end{equation}

This entropy metric quantifies the uncertainty level of the policy on the current training distribution and is widely adopted in maximum entropy RL as a regularization term to encourage exploration~\cite{haarnoja2018soft,ji2024ace,eysenbach2022maximum}.

\begin{figure*}[t]
\centering
\includegraphics[width=1\linewidth]{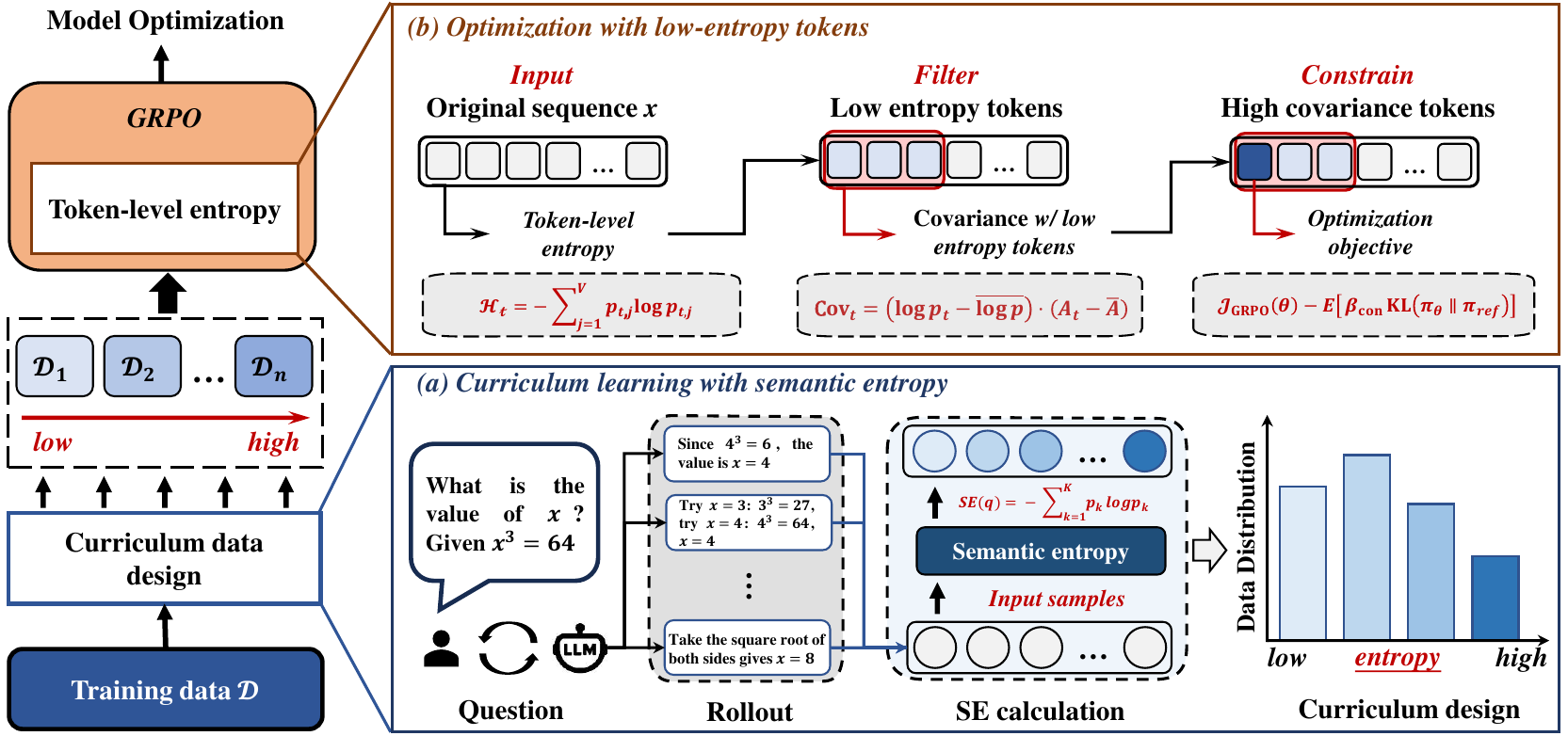}
\caption{Overall framework of \method with two components: (a) Curriculum learning with semantic entropy. (b) Optimization with low-entropy tokens.}
\label{fig:framework}
\end{figure*}

\section{Approach}
\label{sec:approach}

As shown in Fig.~\ref{fig:framework}, \method comprises two components: (1) semantic entropy-guided curriculum learning that computes semantic entropy for each training query and organizes data progressively from low to high entropy for structured difficulty escalation (Section~\ref{sec:curr_learning}), and (2) token-level optimization that identifies low-entropy tokens vulnerable to collapse and applies adaptive KL regularization with stronger constraints on high-covariance portions to encourage exploration (Section~\ref{sec:token_entropy}). Then, we provide the theoretical analysis for preventing entropy collapse (Section~\ref{sec:theory}), and practical implementation of \method (Section~\ref{sec:practical}).

\subsection{Curriculum Learning with Data-Level Semantic Entropy}
\label{sec:curr_learning}

While existing token-level entropy methods address optimization characteristics, they neglect the critical role of training data organization in shaping the global entropy changes. The difficulty distribution of training samples fundamentally influences the reasoning, while a well-structured curriculum can guide the policy to progressively build reasoning capabilities while maintaining efficient exploration.
\textbf{Our key insight is that organizing training data according to their difficulty, measured by semantic entropy that can prevent abrupt entropy drops and facilitate gradual adaptation to increasingly complex reasoning tasks.} This data-level curriculum learning provides a complementary approach to algorithmic optimization, addressing entropy collapse from a global perspective rather than the local token level.

\subsubsection{Semantic Entropy}
Unlike token-level entropy that measures uncertainty in individual token predictions, semantic entropy quantifies the diversity of semantically distinct solutions for a given problem~\cite{farquhar2024detecting,kuhn2023semantic}.

For a query $q$, semantic entropy evaluates whether different generated responses convey the same underlying meaning, even when expressed differently. This is formalized through semantic equivalence classes: responses that are semantically equivalent (meaning the same thing) are grouped together, and the entropy is computed over these meaning clusters rather than individual token sequences. High semantic entropy indicates that the model generates diverse reasoning paths with different solutions, suggesting the problem is challenging and requires extensive exploration. Low semantic entropy indicates the model consistently converges to the same solution, suggesting the problem is relatively easier for the current policy. We compute semantic entropy through a three-step process.

\textbf{Response Generation:}
For each query $q$ in the training dataset $\mathcal{D}$, we sample $M$ responses from the current policy $\pi_{\theta}$:
\begin{equation}
\{o^1, o^2, \ldots, o^M\} \sim \pi_{\theta}(\cdot | q),
\end{equation}
along with their associated probabilities $\{P(o^1|q),\cdots,P(o^M|q) \}$. In \method, we use nucleus sampling with temperature $T=1.0$ to generate diverse responses that reflect the model's uncertainty distribution.

\textbf{Semantic Clustering:}
We cluster the generated responses into semantic equivalence classes based on their meanings. For reasoning tasks with verifiable answers (e.g., mathematics), we operate semantic equivalence through answer equivalence: two responses $o^i$ and $o^j$ are semantically equivalent if they yield the same final answer (e.g., the numerical result in math problems). 
This partitions the $M$ responses into $K$ equivalence classes $\mathcal{C} = \{C_1, C_2, \ldots, C_K\}$, where each class $C_i$ contains responses that express the same meaning.

For each semantic cluster $C_i$, we compute its probability by summing the probabilities of all responses in that cluster:
\begin{equation}
P(C_i | q) = \sum_{o \in C_i} P(o | q).
\end{equation}

\textbf{Entropy Calculation:}
We estimate the semantic entropy as the entropy over the meaning distribution:
\begin{equation}
\mathcal{H}_{\text{SE}}(q) = -\sum_{i=1}^{K} P(C_i | q) \log P(C_i | q),
\label{eq:semantic_entropy}
\end{equation}
to account for sampling limitations, we normalize the cluster probabilities to form a proper probability distribution:
\begin{equation}
\hat{P}(C_i | q) = \frac{P(C_i | q)}{\sum_{j=1}^{K} P(C_j | q)}.
\end{equation}

The final semantic entropy is then computed using these normalized probabilities for Eq.~\ref{eq:semantic_entropy}.

This semantic entropy metric captures the intrinsic difficulty of each query from the perspective of the training data: queries with high semantic entropy exhibit diverse solution distributions and require more exploration, while queries with low semantic entropy show solution convergence and are easier for the model.

\subsubsection{Curriculum Design}

Based on semantic entropy, we design a progressive curriculum that organizes training data from easy to hard. 

Before training, we compute the semantic entropy $\mathcal{H}_{\text{SE}}(q)$ for each query $q \in \mathcal{D}$ using the initial policy $\pi_{\theta}$
(typically the SFT model). This provides a difficulty profile of the entire training dataset from the model's perspective.

Then, we sort the training dataset in ascending order of semantic entropy:
\begin{equation}
\mathcal{D}_{\text{sorted}} = \text{sort}(\mathcal{D}, \text{key}=\mathcal{H}_{\text{SE}}),
\label{eq:cu_data}
\end{equation}
such that queries with lower semantic entropy (easier problems with more consistent solutions) appear earlier, and queries with higher semantic entropy (harder problems requiring diverse exploration) appear later.
Finally, we divide the sorted dataset into $N$ curriculum stages and train the policy progressively. At stage $n \in \{1, 2, \ldots, N\}$, the model is trained on $\mathcal{D}_n$. The selection of stages $N$ is analyzed in section~\ref{sec:curr_analysis}.

This curriculum design prevents the model from encountering overly challenging problems prematurely, which would cause aggressive policy updates and rapid entropy collapse. By building reasoning capabilities on easier examples first, the model maintains stable exploration while gradually adapting to more complex reasoning tasks.

\subsection{Optimization with Token-Level Entropy}
\label{sec:token_entropy}
Most existing methods apply uniform optimization strategies across all token positions, treating each token equally regardless of its impact on policy exploration. This uniform approach has critical shortcomings: it fails to recognize that different tokens play fundamentally different roles in the reasoning process.

To prevent over-optimization at low-entropy positions, we introduce token-selective KL regularization that applies constraints based on token-level entropy. \textbf{Our key insight is that low-entropy tokens are most vulnerable to entropy collapse and require stronger regularization to encourage exploration.}

As demonstrated in~\cite{wang2025beyond}, approximately 80\% of low-entropy tokens significantly influence the learning process for reasoning. However, while that work masks 80\% of low-entropy tokens during optimization, \method takes a different approach: instead of masking all low-entropy tokens which may lead to instability, we apply targeted KL constraints on these tokens to maintain stable exploration while preserving their contribution to learning, with detailed comparison in section~\ref{sec:main_results}. We identify low-entropy tokens as:
\begin{equation}
\mathcal{T}_{\text{low}} = \{o_t \mid \mathcal{H}_t(q, o_{<t}) < \tau_{\mathcal{H}}\},
\label{eq:low_entropy_tokens}
\end{equation}
where $\tau_{\mathcal{H}}$ is the low entropy threshold.

Moreover, not all low-entropy tokens contribute equally to entropy collapse. Following recent analysis~\cite{cui2025entropy}, policy entropy changes are proportional to the covariance between action probabilities and their gradient magnitudes. This suggests that within the low-entropy regime, tokens with high covariance are particularly influential in entropy dynamics.

We compute the covariance for each low-entropy token $o_t \in \mathcal{T}_{\text{low}}$ as:
\begin{equation}
\begin{aligned}
\hspace{-2mm}
 & \text{Cov}_{o_t \sim \pi_{\theta}(\cdot|q,o_{<t})} =\Big(\log{\pi_\theta (o_t|q,o_{<t})}-
\\ & \quad \frac{1}{N} { \sum_{j=1}^{N}}\log
 \pi_{\theta}(o_j|q,o_{<j})   \Big) \cdot (A_t-\frac{1}{N} \sum_{j=1}^{N}A_j),
\label{eq:covariance}
\end{aligned}
\end{equation}
where $N$ is a batch of rollout tokens. For simplicity, we use $\text{Cov}_t$ to represent $ \text{Cov}_{o_t \sim \pi_{\theta}(\cdot|q,o_{<t})}$ in the later sections. This formulation captures the correlation between the model's confidence (log probability) and the learning signal (advantage) at each token. We identify high-covariance portions within low-entropy tokens as:
\begin{equation}
\mathcal{T}_{\text{high-cov}} = \{o_t \in \mathcal{T}_{\text{low}} \mid \text{Cov}_t > \tau_{\text{cov}}\},
\label{eq:high_cov_tokens}
\end{equation}
where $\tau_{\text{cov}}$ is the threshold for high covariance across low-entropy tokens. The selection of thresholds $\tau_{\text{cov}}$ and $\tau_{\mathcal{H}}$ are analyzed in section~\ref{sec:hps}.
Based on token-level entropy and covariance analysis, we apply different KL constraints to tokens according to their vulnerability to entropy collapse. For each token $o_t$, the KL coefficient $\beta_{\text{con}}$ is determined by:
\begin{equation}
\beta_{\text{con}} =
\begin{cases}
\beta_{\text{low}} & \text{if } o_t \in \mathcal{T}_{\text{low}} \setminus \mathcal{T}_{\text{high-cov}}, \\
\beta_{\text{high}} & \text{if } o_t \in \mathcal{T}_{\text{high-cov}} \subseteq \mathcal{T}_{\text{low}}, \\
0 & \text{if } o_t \notin \mathcal{T}_{\text{low}},
\end{cases}.
\label{eq:adaptive_beta}
\end{equation}

This fine-grained constraint mechanism ensures that regularization strength is proportional to the token's influence on entropy changes.

\textbf{Optimization Objective:}
Combining the data-level curriculum with token-level selective regularization, the optimization objective of \method is:
\begin{equation}
\hspace{-5mm}
\begin{aligned}
\mathcal{J}_{\method}(\theta)
&= \mathbb{E}_{(q,a)\sim \mathcal{D}_n,\,\{o^i\}_{i=1}^G \sim \pi_{\mathrm{old}}(\cdot | q)}
\Bigg[ \frac{1}{G} \sum_{i=1}^{G} \frac{1}{|o^i|} \sum_{t=1}^{|o^i|}  \\
& \quad \bigg(\min\!\big(r^i_t(\theta)\hat{A}^i,\, \mathrm{clip}(r^i_t(\theta), 1-\epsilon, 1+\epsilon)\hat{A}^i\big) \\
& \quad\quad - \beta_{\text{con}} \, D_{\mathrm{KL}}(\pi_{\theta} \,\|\, \pi_{\mathrm{ref}}) \bigg)
\Bigg],
\end{aligned}
\label{eq:final_objective}
\end{equation}
where $\mathcal{D}_n$ is the curriculum-stage dataset from Eq.~\ref{eq:cu_data}, $\beta_{\text{con}}$ is the KL coefficient from Eq.~\ref{eq:adaptive_beta}, and other terms follow the standard GRPO formulation.

\subsection{Theoretical Analysis}
\label{sec:theory}

In this section, we provide theoretical analysis for the entropy changes of our \method framework. Building upon~\cite{cui2025entropy}, we show how our token-selective KL regularization controls entropy decline and maintains exploration capacity.

\subsubsection{Entropy Changes}

Following the theoretical framework in~\cite{cui2025entropy}, we establish how policy entropy evolves under our token-level optimization. For softmax policies like LLMs, the entropy change between consecutive training steps is determined by the covariance between action log-probabilities and logit changes.

\textbf{Lemma 1} (Entropy changes~\cite{cui2025entropy,liu2025does}). \textit{For a softmax policy $\pi_{\theta}$, the entropy change given state $s$ between two consecutive steps $k$ and $k+1$ under first-order approximation satisfies:}
\begin{equation}
\begin{aligned}
& \mathcal{H}(\pi^{k+1}_{\theta}) - \mathcal{H}(\pi^{k}_{\theta})  \\ & \quad \approx - \mathbb{E}_{s_t} \Big[ \text{Cov}_t \left(\log \pi^k_{\theta}(o_t|s_t), \theta^{k+1}_{s_t,o_t} - \theta^k_{s_t,o_t}\right) \Big],
\end{aligned}
\label{eq:entropy_change}
\end{equation}
\textit{where $s_t=(q,o_{<t})$ denotes the state at position $t$, and $\theta_{s_t,o_t}$ denotes the output logit of $(s_t,o_t)$. $\theta^{k+1}_{s_t,o_t} - \theta^k_{s_t,o_t}$ is the change in the output logits between step $k$ and step $k + 1$. It indicates that the change of policy entropy approximately equals the negative covariance between log-probability of the action and the change of logits.}

Under Policy Gradient algorithms like GRPO, the logit change is:
\begin{equation}
\theta^{k+1}_{s,o} - \theta^k_{s,o} = \eta\cdot\nabla_\theta\mathcal{J}(\theta).
\end{equation}

Meanwhile, it is a common practice to use the Policy Gradient algorithm~\cite{williams1992simple} for gradient estimation:
\begin{equation}
\hspace{-3mm}
\begin{aligned}
\nabla\mathcal{J}(\theta)
&= \mathbb{E}_{s\sim \mathcal{D}_n,\, o_t \sim \pi_{\theta}(\cdot | s)}
\Bigg[ \sum_{t=0}^{T}\nabla_{\theta}\log \pi_{\theta}(o_t|s)A_t
\Bigg].
\end{aligned}
\end{equation}

\textbf{Proposition 1} (Logit Change in Policy Gradient~\cite{NIPS2001_4b86abe4,cui2025entropy}). \textit{When updating policy via Policy Gradient with learning rate $\eta$, the logit difference satisfies:}
\begin{equation}
\theta^{k+1}_{s_t,o_t} - \theta^k_{s_t,o_t} = \eta\cdot\pi^k_\theta(o_t|s_t)A_t.
\label{eq:logit_difference}
\end{equation}

Combining Lemma 1 and Proposition 1, we obtain:

\textbf{Theorem 1} (Entropy Change under Vanilla Policy Gradient~\cite{liu2025does,cui2025entropy}). \textit{For Policy Gradient updates, the entropy change satisfies:}
\begin{equation}
\hspace{-4mm}
\begin{aligned}
& \mathcal{H}(\pi^{k+1}_{\theta}) - \mathcal{H}(\pi^{k}_{\theta})\\
&\quad  \approx -\eta \, \mathbb{E}_{s_t} \Big[\text{Cov}_t\big(\log \pi^k_{\theta}(o_t|s_t), \pi^k_{\theta}(o_t|s_t)A_t\big) \Big].
\label{eq:entropy_vanilla}
\end{aligned}
\end{equation}

\subsubsection{Token-Level KL Regularization}

Now we analyze how our token-level KL regularization affects entropy changes. The optimization objective $\mathcal{J}_\method$ applies differentiated KL coefficients $\beta_{\text{con}}$.
By projecting the gradient to the logit space and taking a first-order approximation, the logit update is:
\begin{equation}
\hspace{-4mm}
\theta^{k+1}_{s_t,o_t} - \theta^k_{s_t,o_t}=
\eta\,\Big(\pi_\theta^k(o_t| s_t) A_t
-\beta_{\text{con}}\,
\nabla_\theta D_{\mathrm{KL}}(\pi_\theta^k\|\pi_{\mathrm{ref}})\Big).
\label{eq:L2}
\end{equation}

We provide detailed derivation of Eq.~\ref{eq:entropy_vanilla} and~\ref{eq:L2} in the Appendix.

\textbf{Theorem 2} (Entropy Changes with Token-Level KL Regularization). \textit{By combining Eq.~\ref{eq:L2} into Eq.~\ref{eq:entropy_change}, we can get the entropy change based on Theorem 1 for state $s_t$ satisfies:}
\begin{equation}
\hspace{-5mm}
\begin{aligned}
& \mathcal{H}(\pi^{k+1}_{\theta}) - \mathcal{H}(\pi^{k}_{\theta}) \\ & \quad \approx -  \mathbb{E}_{s_t} \Big[\text{Cov}_t\big(\log \pi^k_{\theta}(o_t|s_t), \theta^{k+1}_{s_t,o_t} - \theta^k_{s_t,o_t}\big) \Big]
\\ & \quad = - \mathbb{E}_{s_t} \Big[\text{Cov}_t\Big(\log \pi^k_{\theta}(o_t|s_t), 
\eta\cdot\nabla_\theta\mathcal{J}_\method(\theta)\Big) \Big]
\\ & \quad = \underbrace{ -\eta\mathbb{E}_{s_t} \Big[\text{Cov}_{t}\left(\log \pi^k_{\theta}(o_t|s_t),\pi^k_{\theta}(o_t|s_t)A_t\right) \Big]}_{\text{Term 1}}  +
\\ & \quad\quad  \underbrace{\eta\mathbb{E}_{s_t} \Big[\beta_{\mathrm{con}} \,\text{Cov}_{t}\left(\log \pi^k_{\theta}(o_t|s_t), \nabla_\theta D_{\mathrm{KL}}(\pi_\theta^k\|\pi_{\mathrm{ref}})\right) \Big]}_{\text{Term 2}}.
\end{aligned}
\label{eq:entropy_token_selective}
\end{equation}

Eq.~\ref{eq:entropy_token_selective} reveals how our token-selective KL regularization controls entropy dynamics. The entropy change consists of two terms:

\textbf{Term 1: Vanilla Entropy Decay.} The first term $ -\eta\mathbb{E}_{s_t} \Big[\text{Cov}_{t}\left(\log \pi^k_{\theta}(o_t|s_t),\pi^k_{\theta}(o_t|s_t)A_t\right) \Big]$ is the standard entropy change from Policy Gradient, which typically drives entropy collapse. This covariance term is predominantly positive during training (high-probability tokens with high advantages), leading to monotonic entropy decrease.

\textbf{Term 2: KL-Induced Entropy Preservation.} The second term $\eta\mathbb{E}_{s_t} \Big[\beta_{\mathrm{con}} \,\text{Cov}_{t}\left(\log \pi^k_{\theta}(o_t|s_t), \nabla_\theta D_{\mathrm{KL}}(\pi_\theta^k\|\pi_{\mathrm{ref}})\right) \Big]$ represents the entropy-preserving effect of our KL regularization. The KL term acts to pull the policy back toward the reference distribution, reducing the magnitude of policy updates. Critically, this term has a \textit{positive contribution} to entropy change, counteracting the entropy collapse from Term 1. 
Our hierarchical KL coefficient $\beta_{\text{con}}$ provides differentiated entropy control across token types:

\textit{Case 1: High-Entropy Tokens ($o_t \notin \mathcal{T}_{\text{low}}$).} For tokens with $\mathcal{H}_t > \tau_{\mathcal{H}}$, we set $\beta_{\text{con}} = 0$. These tokens already exhibit sufficient exploration diversity, so no additional regularization is needed. The entropy change follows the vanilla Policy Gradient.

\textit{Case 2: Low-Entropy Tokens ($o_t \in \mathcal{T}_{\text{low}} \setminus \mathcal{T}_{\text{high-cov}}$).} For tokens with $\mathcal{H}_t < \tau_{\mathcal{H}}$ but moderate covariance, we apply $\beta_{\text{con}} = \beta_{\text{low}}$. These near-deterministic tokens are vulnerable to over-optimization. The moderate KL penalty slows entropy decline.

\textit{Case 3: High-Covariance Low-Entropy Tokens ($o_t \in \mathcal{T}_{\text{high-cov}} \subseteq \mathcal{T}_{\text{low}}$).} For tokens with both low entropy ($\mathcal{H}_t < \tau_{\mathcal{H}}$) and high covariance ($\text{Cov}_t > \tau_{\text{cov}}$), we apply the strongest KL regularization $\beta_{\text{con}} = \beta_{\text{high}} $. According to Eq.~\ref{eq:entropy_vanilla}, high covariance indicates these tokens are most influential for entropy collapse. The strong KL penalty provides maximal entropy preservation.

\textbf{Entropy Preservation.} Since $\beta_{\text{high}} > \beta_{\text{low}} > 0$, the KL-induced preservation term is strongest where entropy collapse is most severe. This hierarchical design ensures that:
\begin{itemize}
\item \textbf{Targeted Intervention:} Only vulnerable tokens (low-entropy) receive regularization, avoiding unnecessary constraints on exploratory (high-entropy) tokens.
\item \textbf{Covariance-Aware Prioritization:} Within low-entropy tokens, those with high covariance identified as primary drivers of entropy collapse receives strongest constraints.
\item \textbf{Balanced Optimization:} The interplay between entropy decay (Term 1) and preservation (Term 2) maintains $\mathcal{H}(\pi_{\theta}) \geq \mathcal{H}_{\min} > 0$ throughout training, unlike vanilla where $\mathcal{H}(\pi_{\theta}) \to 0$.
\end{itemize}

Unlike all-tokens entropy regularization (which applies equal penalties to all tokens) or uniform covariance methods (which treat all high-covariance tokens equally), \method provides \textit{fine-grained control}: we only intervene where entropy collapse is most likely (low-entropy) and most impactful (high-covariance), achieving entropy preservation with minimal interference to the optimization process.

\textbf{Combined with Curriculum Learning.} The data-level curriculum complements this token-level control by organizing training samples from low to high semantic entropy. Easy samples (low semantic entropy) allow the model to identify stable low-entropy tokens that can safely converge, while hard samples (high semantic entropy) require maintaining diversity at more positions. Our token-level regularization adapts to this curriculum progression, providing dynamic entropy control throughout training. The curriculum component adds an additional stabilization mechanism absent in existing works, preventing entropy collapse from a complementary data-organization perspective.

\subsection{Practical Implementation}
\label{sec:practical}

\begin{algorithm}[t]
\caption{\method}
\label{alg:method}
\begin{algorithmic}[1]
\STATE \textbf{Input:} Training dataset $\mathcal{D}$, initial policy $\pi_\theta$, curriculum stages $N$, sampling number $M$, thresholds $\tau_{\mathcal{H}}$, $\tau_{\text{cov}}$
\STATE \textbf{Output:} Optimized policy $\pi_\theta$
\item[] {\color{darkred} // \textbf{Curriculum design with semantic entropy}}
\FOR{each query $q$ in $\mathcal{D}$}
    \STATE Sample $M$ responses $\{o^1,\ldots,o^M\}$ from $\pi_\theta(\cdot|q)$
    \STATE $\mathcal{C} = \{C_1,\ldots,C_K\}$ 
    \STATE $\mathcal{H}_{\text{SE}}(q) = -\sum_i^{K} P(C_i|q)\log P(C_i|q)$
\ENDFOR
\STATE Sort $\mathcal{D}$ by ascending $\mathcal{H}_{\text{SE}}(q)$
\item[] {\color{darkred} // \textbf{Optimization with low-entropy tokens}}
\FOR{$n=1$ to $N$}
        \STATE Initialize empty set $\mathcal{T}_{\text{low}}$, $\mathcal{T}_{\text{high-cov}}$
    \FOR{each $(q,a)\in\mathcal{D}_n$}
        \FOR{each token $o_t$ in response sequence}
            \STATE Compute token entropy $\mathcal{H}_t(q,o_{<t})$
            \IF{$\mathcal{H}_t < \tau_{\mathcal{H}}$}
                \STATE Add $o_t$ to $\mathcal{T}_{\text{low}}$ \hspace{1mm} $\triangleright$ \texttt{Low entropy tokens}
            \ENDIF
        \ENDFOR

        \FOR{each $o_t$ in $\mathcal{T}_{\text{low}}$}
            \STATE Compute covariance $\text{Cov}_t$ following Eq.~\ref{eq:covariance}
            \IF{$\text{Cov}_t > \tau_{\text{cov}}$}
                \STATE Add $o_t$ to $\mathcal{T}_{\text{high-cov}}$  \hspace{1mm} $\triangleright$ \texttt{High covariances}
            \ENDIF
        \ENDFOR

        \FOR{each token $o_t$ in response sequence}
            \STATE Set $\beta_{\text{con}}$ following Eq.~\ref{eq:adaptive_beta} \hspace{0.03mm} $\triangleright$ \texttt{KL coefficient} 
            \STATE Update parameters $\theta$ following Eq.~\ref{eq:final_objective}
        \ENDFOR
    \ENDFOR
\ENDFOR

\end{algorithmic}
\end{algorithm}

Algorithm~\ref{alg:method} presents the complete training procedure for our \method framework. The algorithm integrates semantic entropy-based curriculum learning at the data level with token-selective KL regularization at the algorithmic level, achieving synergistic exploration-aware optimization. 
First, semantic entropy is computed for each query to quantify data difficulty, and the training dataset is sorted accordingly to form a progressive curriculum from easy to hard  (lines 3-8). During training, token-level entropy is used to identify low-uncertainty tokens that are prone to collapse  (lines 12-17). Covariance between token probabilities and advantages is further evaluated to detect highly influential tokens within low entropy tokens (lines 18-23). Based on entropy and covariance, adaptive KL regularization with coefficient $\beta_{\text{con}}$ is applied, imposing stronger constraints on stable but high-covariance tokens while allowing flexible updates for others  (lines 24-27). This joint data-level and token-level mechanism enables stable exploration and reasoning enhancement.

\section{Experiment}
\label{sec:experiments}
Our experiments aim to address the following questions: (i) How does the performance of \method compare to other entropy-guided approaches in diverse tasks. (ii) Can \method, through curriculum learning and token-level entropy optimization, mitigating entropy collapse and improving LLMs' reasoning? (iii) What are the effects of the components and hyperparameters in \method? (iv) What is the generalization performance in out-of-distribution tasks? (v) What insights can be gained from case studies of specific dialogues? 

\subsection{Experimental Setup}
\label{sec:setup}

\subsubsection{Benchmarks and Implementation Details} We conduct training on different-sized base models, including: DeepSeek-R1-Distill-Qwen-1.5B, Qwen2.5-Math-7B, and Qwen3-14B on DAPO-MATH-17K datasets~\cite{yu2025dapo}. The rollout size $M$ is $8$, temperature factor is $1.0$, max response length is $2048$, curriculum stage is $2$, $\beta_{\rm{high}}$ is $2$, $\beta_{\rm{low}}$ is $0.5$, and learning rate $\eta$ is $1e-6$. For fair comparisons, we reproduce all baselines and conduct training under the same hyperparameters in the VeRL platform~\cite{sheng2025hybridflow}. We select GRPO as our basic post-training algorithm for applying entropy-guided methods. We conduct extensive validation on six benchmarks, including AIME 2025\&2024~\cite{codeforcesamerican}, AMC 2023~\cite{amc23}, MATH500~\cite{hendrycks2measuring}, OlympiadBench~\cite{he2024olympiadbench}, and Minerva~\cite{lewkowycz2022solving}. We provide more experimental details in the Appendix.

\subsubsection{Baselines} We conduct comparison with baselines including directly adding entropy into the learning objective~\cite{haarnoja2018soft} (w/ En), adding an entropy-based advantage function~\cite{cheng2025reasoning} (w/ Adv), masking low-entropy tokens~\cite{wang2025beyond} (w/ Mask), clipping a small fraction of high-covariance tokens~\cite{cui2025entropy} (w/ Clip) and constrain high covariance tokens~\cite{cui2025entropy} (w/ Cov). Moreover, we also compare with adding a high-entropy reward for optimization (w/ High\_En) to validate the advantages of constraining low-entropy tokens over encouraging high-entropy tokens.

\subsubsection{Metrics} We assess the reasoning ability boundaries using the \textit{Pass@K}: represents at least one of $K$ sampled model outputs passes verification, and assess the average performance using the \textit{Avg@K}: denotes the average accuracy over $K$ evaluations, \textit{Len@K}: average response length over $K$ evaluations per benchmark. We vary $K$ across experiments to ensure the statistical reliability of our evaluation results.

\subsection{Main Results}
\label{sec:main_results}

\begin{table*}[t]
 \caption{Performance comparison on six datasets under DeepSeek-R1-Distill-Qwen-1.5B base model. w/ means with. "Pass@k" represents at least one of
$K$ sampled model outputs passes verification. We bold the best scores and underline the sub-optimal results.} 
\renewcommand{\arraystretch}{1.2}
\setlength{\tabcolsep}{3.2pt} 
    \centering 
\begin{tabular}{ccccccccccc}
\toprule
\rowcolor{black!10}   \textbf{Benchmark} &  \textbf{Metric}                & \textbf{Base model} & \textbf{GRPO}~\cite{guo2025deepseek} & \textbf{w/ En}~\cite{haarnoja2018soft} & \textbf{w/ Adv}~\cite{cheng2025reasoning} & \textbf{w/ Mask}~\cite{wang2025beyond} & \textbf{w/ Clip}~\cite{cui2025entropy} & \textbf{w/ Cov}~\cite{cui2025entropy}  & \textbf{w/ High\_En} & \textbf{w/ \method} \\ \bottomrule
\multirow{3}{*}{\textbf{AIME24}}        & \textit{Pass@8}  & 40.00               & 40.00         & \underline{60.00}               & 53.33                         & 56.67    & 46.67         & 43.33               & 46.67                            & \textbf{63.33}          \\ 
                                        & \textit{Pass@16} & 43.33                & 50.00          & 60.00               & 56.67                         & \underline{63.33} & 46.67        & 53.33               & 60.00                            & \textbf{70.00}          \\
                                        & \textit{Pass@32} & 43.33                & 56.67         & 70.00               & 53.33                         & \underline{73.33} & 70.00            & 53.33               & 53.33                            & \textbf{76.67}          \\ \hline
\multirow{3}{*}{\textbf{AIME25}}        & \textit{Pass@8}  & 26.67                & 33.33          & 33.33               & 33.33                         & 33.33     & 33.33        & \underline{36.67}               & 33.33                            & \textbf{40.00}          \\
                                        & \textit{Pass@16} & 33.33                & \underline{40.00}          & 36.67               & 30.00                         & 33.33  & \underline{40.00}           & 36.67               & 33.33                            & \textbf{43.33}          \\
                                        & \textit{Pass@32} & 30.00                & \underline{43.33}          & \underline{43.3}               & 36.67                         & \underline{43.33} & 36.67             & 40.00               & 40.00                            & \textbf{46.67}          \\ \hline
\multirow{3}{*}{\textbf{AMC23}}         & \textit{Pass@8}  & 80.00                & 87.50          & \underline{90.00}               & \underline{90.00}                         & \underline{90.00}             & \textbf{92.50} & \textbf{92.50}               & 85.00                            & \textbf{92.50}          \\
                                        & \textit{Pass@16} & 87.50                & \textbf{95.00}          & 90.00               & \underline{92.50}                         & \underline{92.50}             & \underline{92.50}             & 90.00               & \underline{92.50}                            & \textbf{95.00}          \\
                                        & \textit{Pass@32} & 92.50                & \underline{95.00}          & \textbf{97.50}               & \underline{95.00}                         & \underline{95.00}             & 90.00             & \textbf{97.50}               & \underline{95.00}                            & \textbf{97.50}          \\ \hline
\multirow{3}{*}{\textbf{MATH500}}       & \textit{Pass@8}  & 90.00                & 91.60          & 91.60               & 89.60                         & \underline{92.80}   & \textbf{93.00}           & 92.60               & 91.00                            & \textbf{93.00}          \\
                                        & \textit{Pass@16} & 92.40                & \textbf{94.00}          & 93.20               & 91.60                         & 93.60             & 93.00             & \underline{93.80}               & 92.80                            & \underline{93.80}          \\
                                        & \textit{Pass@32} & 93.20                & \underline{95.00}          & 94.60               & 94.00                         & 94.40 & 93.80             & 94.80               & 94.20                            & \textbf{95.20}          \\ \hline
\multirow{3}{*}{\textbf{OlympiadBench}} & \textit{Pass@8}  & 52.89                & 54.96          & \underline{60.89}               & 54.81                         & 60.15 & 58.96             & 56.44               & 58.37                            & \textbf{63.26}          \\
                                        & \textit{Pass@16} & 58.52                & 59.41          & 62.67               & 60.00                         & \underline{65.19} & 64.89             & 64.15               & 63.85                            & \textbf{67.26}          \\
                                        & \textit{Pass@32} & 61.19               & 64.30          & \underline{69.33}               & 63.41                         & 69.19 & 68.74             & 67.70               & 66.67                            & \textbf{70.22}          \\ \hline
\multirow{3}{*}{\textbf{Minerva}}       & \textit{Pass@8}  & 33.82                & 32.72          & 34.56               & 34.56                         & 34.93       & 33.09      & \underline{36.76}               & 34.56                            & \textbf{37.50}          \\
                                        & \textit{Pass@16} & 36.76                & 36.03          & \underline{41.91}               & 37.13                         & 38.97 & 38.60             & 40.44               & 41.18                            & \textbf{42.65}          \\
                                        & \textit{Pass@32} & 41.54                & 39.71          & \underline{45.59}               & 40.44                         & \underline{45.59} & 44.49             & 43.01               & 45.22                            & \textbf{46.32}          \\ \hline
                                        \rowcolor{black!10} \multicolumn{2}{c}{\textbf{Avg.}} &  57.61 & 61.59 & 65.29 & 61.47 & \underline{65.31} & 63.16 & 62.95 & 62.61 & \textbf{68.57} \\ \bottomrule
\end{tabular}
\label{tab:main_res_pass}
\end{table*}

\begin{table*}[t]
 \caption{Performance comparison on six datasets under DeepSeek-R1-Distill-Qwen-1.5B base models. w/ means with. "Avg@k" denotes the average accuracy over $K$ evaluations per benchmark. We bold the best scores and underline the sub-optimal results.} 
\renewcommand{\arraystretch}{1.2}
\setlength{\tabcolsep}{3.2pt} 
    \centering 
\begin{tabular}{ccccccccccc}
\toprule
\rowcolor{blue!10}    \textbf{Benchmark} &  \textbf{Metric} & \textbf{Base model} & \textbf{GRPO}~\cite{guo2025deepseek} & \textbf{w/ En}~\cite{haarnoja2018soft} & \textbf{w/ Adv}~\cite{cheng2025reasoning} & \textbf{w/ Mask}~\cite{wang2025beyond} & \textbf{w/ Clip}~\cite{cui2025entropy} & \textbf{w/ Cov}~\cite{cui2025entropy} & \textbf{w/ High\_En} & \textbf{w/ \method} \\ \bottomrule
\multirow{3}{*}{\textbf{AIME24}}        & \textit{Avg@8}  & 18.33                & 14.17          & 22.50               & 21.67                         & \underline{23.33} & \underline{23.33}            & 19.17               & 19.58                            & \textbf{27.92}          \\ 
                                        & \textit{Avg@16} & 18.96                & 16.46          & 20.62               & 19.58                         & \underline{24.79} & 22.50             & 21.67               & 21.25                            & \textbf{25.21}          \\
                                        & \textit{Avg@32} & 18.54               & 15.63         & 22.29               & 19.58                         & \underline{24.79} & 24.27             & 20.52               & 18.96                            & \textbf{26.56}          \\ \hline
\multirow{3}{*}{\textbf{AIME25}}        & \textit{Avg@8}  & 19.17                & 15.83          & 16.67               & 18.75                         & \underline{20.83} & \underline{20.83}            & 19.17               & 19.17                           & \textbf{21.25}          \\
                                        & \textit{Avg@16} & \underline{19.17}                & 15.63          & 17.08               & 17.92                         & 17.92 & 18.54             & 17.08               & 17.71                            & \textbf{21.46}          \\
                                        & \textit{Avg@32} & 17.71                & 13.75          & 18.12               & 18.33                         & 19.38 & \underline{19.79}             & 19.48               & 17.08                            & \textbf{20.10}          \\ \hline
\multirow{3}{*}{\textbf{AMC23}}         & \textit{Avg@8}  & 57.81                & 56.25          & 65.62               & 60.31                         & \underline{70.00} & 64.49             & 60.94               & 60.00                            & \textbf{71.56}          \\
                                        & \textit{Avg@16} & 57.03                & 60.47          & 63.75               & 56.41                         & \underline{68.28} & 67.03             & 65.94               & 62.97                            & \textbf{70.16}          \\
                                        & \textit{Avg@32} & 57.58                & 58.12          & 65.47               & 58.12                         & \underline{69.61} & 65.23             & 64.38               & 64.69                            & \textbf{70.00}          \\ \hline
\multirow{3}{*}{\textbf{MATH500}}       & \textit{Avg@8}  & 77.17                & 73.88          & 79.40               & 77.02                         & 80.60             & \underline{80.95} & 80.35               & 78.85                            & \textbf{81.48}          \\
                                        & \textit{Avg@16} & 77.26                & 73.47          & 79.98               & 76.60                         & \underline{80.76} & 80.24             & 80.31               & 78.86                            & \textbf{81.74}          \\
                                        & \textit{Avg@32} & 76.54                & 73.78          & 79.69               & 76.83                         & \textbf{81.37} & 80.94            & 79.89               & 79.57                            & \underline{81.22}          \\ \hline
\multirow{3}{*}{\textbf{OlympiadBench}} & \textit{Avg@8}  & 35.65                & 29.48          & 40.48               & 37.04                         & \underline{42.65}  & 41.44            & 39.72               & 39.59                            & \textbf{42.85}          \\
                                        & \textit{Avg@16} & 36.59                & 29.43          & 40.11               & 36.28                         & \underline{42.89} & 41.36             & 40.29               & 39.70                            & \textbf{42.99}          \\
                                        & \textit{Avg@32} & 36.51                & 29.60          & 40.39               & 36.24                         & \underline{42.88} & 41.25             & 40.92               & 39.48                            & \textbf{42.95}          \\ \hline
\multirow{3}{*}{\textbf{Minerva}}       & \textit{Avg@8}  & 18.38                & 17.65          & 20.13               & 19.03                         & \underline{21.14} & 20.27             & 20.40               & 19.62                           & \textbf{21.90}          \\
                                        & \textit{Avg@16} & 18.22                & 16.93          & 20.86               & 18.36                         & \underline{21.25} & 20.96             & 20.63               & 20.38                            & \textbf{21.25}          \\
                                        & \textit{Avg@32} & 18.89                & 16.97          & 20.98               & 18.84                         & \underline{21.19} & 20.66             & 20.62               & 20.38                            & \textbf{21.53}          \\ \hline
                                        \rowcolor{blue!10} \multicolumn{2}{c}{\textbf{Avg.}} & 37.75 & 34.86 & 40.79 & 38.16 & \underline{42.98} & 41.89 & 40.64 & 39.88 & \textbf{44.01} \\
                                        \bottomrule
\end{tabular}
\label{tab:main_res_avg}
\end{table*}

\begin{table*}[t]
 \caption{Comparison on six benchmarks under \textit{Pass@16} and \textit{Avg@16} using 7B base models. w/ means with. We bold the best results and underline the sub-optimal results. $\Delta$ means the difference between the results of \method and sub-optimal results.} 
\renewcommand{\arraystretch}{1.2}
\setlength{\tabcolsep}{3.8pt} 
    \centering 
\begin{tabular}{lcccccccccccc}
\toprule
\textbf{}                          & \multicolumn{2}{c}{\textbf{AIME24}} & \multicolumn{2}{c}{\textbf{AIME25}} & \multicolumn{2}{c}{\textbf{AMC23}} & \multicolumn{2}{c}{\textbf{MATH500}} & \multicolumn{2}{c}{\textbf{OlympiadBench}} & \multicolumn{2}{c}{\textbf{Minerva}} \\ \cmidrule(lr){2-3} \cmidrule(lr){4-5} \cmidrule(lr){6-7} \cmidrule(lr){8-9} \cmidrule(lr){10-11} \cmidrule(lr){12-13}
\textbf{Method}                    & \textit{Pass@16}  & \textit{Avg@16} & \textit{Pass@16}  & \textit{Avg@16} & \textit{Pass@16} & \textit{Avg@16} & \textit{Pass@16}  & \textit{Avg@16}  & \textit{Pass@16}     & \textit{Avg@16}     & \textit{Pass@16}  & \textit{Avg@16}  \\ \bottomrule
\rowcolor{black!10} \multicolumn{13}{c}{\textit{Qwen2.5-Math-7B}}  \\ \hline
 \textbf{GRPO}      & 56.67 & 24.38 & 30.00 & 11.25 & 85.00 & 62.50 & \underline{91.80} & \underline{74.76} & 61.04 & 37.56 & \underline{39.71} & 21.51                        \\
 \quad \textbf{w/ En} & \underline{60.00} & \underline{25.42} & \underline{36.67} & \underline{11.67} & \underline{90.00} & \underline{66.56} & 91.00 & \textbf{75.07} & \underline{61.17} & 37.33 & 38.24 & 21.62        \\
\quad \textbf{w/ Mask} & 56.67 & 24.17 & 26.60 & 9.79 & 87.50 & 64.69 & 91.00 & 74.49 & 60.89 & \underline{37.61} & \underline{39.71} & \textbf{22.04}                      \\  \rowcolor{blue!10} 
\quad \textbf{w/ \method} & \textbf{66.67} & \textbf{30.63} & \textbf{50.00} & \textbf{19.79} & \textbf{95.00} & \textbf{69.22} & \textbf{92.00} & 73.83 & \textbf{67.11} & \textbf{38.00} & \textbf{40.07} & \underline{21.76}  \\
\quad\quad $\Delta$ & \color{darkgreen}{+6.67} & \color{darkgreen}{+5.21}  & \color{darkgreen}{+13.33}  & \color{darkgreen}{+8.12} & \color{darkgreen}{+5.00} & \color{darkgreen}{+2.66} & \color{darkgreen}{+0.20} & \color{darkred}{-1.24} & \color{darkgreen}{+5.94} & \color{darkgreen}{+0.39} & \color{darkgreen}{+0.36} & \color{darkred}{-0.28}
\\ \bottomrule 
\end{tabular}
\label{tab:res_larger_model}
\end{table*}

\begin{table*}[t]
 \caption{Comparison on six benchmarks under \textit{Pass@16} and \textit{Avg@16} using 14B base models. w/ means with. We bold the best results. $\Delta$ means the difference between the results of \method and GRPO results.} 
\renewcommand{\arraystretch}{1.2}
\setlength{\tabcolsep}{3.8pt} 
    \centering 
\begin{tabular}{lcccccccccccc}
\toprule
\textbf{}                          & \multicolumn{2}{c}{\textbf{AIME24}} & \multicolumn{2}{c}{\textbf{AIME25}} & \multicolumn{2}{c}{\textbf{AMC23}} & \multicolumn{2}{c}{\textbf{MATH500}} & \multicolumn{2}{c}{\textbf{OlympiadBench}} & \multicolumn{2}{c}{\textbf{Minerva}} \\ \cmidrule(lr){2-3} \cmidrule(lr){4-5} \cmidrule(lr){6-7} \cmidrule(lr){8-9} \cmidrule(lr){10-11} \cmidrule(lr){12-13}
\textbf{Method}                    & \textit{Pass@16}  & \textit{Avg@16} & \textit{Pass@16}  & \textit{Avg@16} & \textit{Pass@16} & \textit{Avg@16} & \textit{Pass@16}  & \textit{Avg@16}  & \textit{Pass@16}     & \textit{Avg@16}     & \textit{Pass@16}  & \textit{Avg@16}  \\ \bottomrule
\rowcolor{black!10} \multicolumn{13}{c}{\textit{Qwen3-14B}}  \\ \hline
 \textbf{GRPO}      & 80.00 & 57.50 & 63.33 & 37.50 & \textbf{100} & 90.16 & 95.60 & 90.39 & 70.37 & 58.39 & 44.12 & 35.02 
                   \\  \rowcolor{blue!10} 
\quad \textbf{w/ \method} & \textbf{86.67} & \textbf{65.62} & \textbf{83.33} & \textbf{48.96} & \textbf{100} & \textbf{93.75} & \textbf{100} & \textbf{93.75} & \textbf{73.33} & \textbf{61.46} & \textbf{45.22} & \textbf{35.34}  \\
\quad\quad $\Delta$ & \color{darkgreen}{+6.67} & \color{darkgreen}{+8.12}  & \color{darkgreen}{+20.00}  & \color{darkgreen}{+11.46} & \color{darkgreen}{+0.00} & \color{darkgreen}{+3.59} & \color{darkgreen}{+4.40} & \color{darkgreen}{+3.36} & \color{darkgreen}{+2.66} & \color{darkgreen}{+3.07} & \color{darkgreen}{+1.10} & \color{darkgreen}{+0.32}
\\ \bottomrule 
\end{tabular}
\label{tab:res_larger_model_qwen3}
\end{table*}

\begin{table*}[t]
 \caption{Performance comparison on six benchmarks on \textit{Len@16}. w/ means with. We bold the best results. $\Delta$ means the difference between the results of \method and sub-optimal results.} 
\renewcommand{\arraystretch}{1.2}
\setlength{\tabcolsep}{8pt} 
    \centering 
\begin{tabular}{llccccccc}
\toprule & \textbf{Method} & \textbf{AMIE24} & \textbf{AIME25} & \textbf{AMC23}  & \textbf{MATH500} & \textbf{OlympiadBench} & \textbf{Minerva}                       & \textbf{Avg.} \\ \bottomrule 
\multirow{4}{*}{\makecell{\textbf{DeepSeek-R1-Distill}\\\textbf{Qwen-1.5B}}} &
\textbf{GRPO}                      & 4007.04 & 5464.50 & 2713.26 & 1410.54  &  1007.16             & 575.82 &   2529.72   \\ &
\quad \textbf{w/ En}                & 4217.96 & 4186.46 & 2974.16 & 2065.79  &   2944.22 & 2266.90                        &  3109.25    \\   &  
\quad \cellcolor{blue!10}{\textbf{w/ \method}} & \cellcolor{blue!10}\textbf{6020.88} & \cellcolor{blue!10}\textbf{5854.60} & \cellcolor{blue!10}\textbf{3707.48} & \cellcolor{blue!10}\textbf{2376.65}  & \cellcolor{blue!10}\textbf{4006.52}        & \cellcolor{blue!10}\textbf{2712.79}                        &    \cellcolor{blue!10}\textbf{4113.15}  \\ &
\quad\quad $\Delta$ & \color{darkgreen}{+1802.92} & \color{darkgreen}{+390.10} & \color{darkgreen}{+733.32} & \color{darkgreen}{+310.96} & \color{darkgreen}{+1062.30} & \color{darkgreen}{+445.89} & \color{darkgreen}{+1003.85}
\\ \hline 
\multirow{4}{*}{\textbf{Qwen2.5-Math-7B}} & \textbf{GRPO}                      &    1335.82    &    1282.48    &   887.77     &    712.33     &      912.59         & 749.84                              &  980.14    \\ &
\quad \textbf{w/ En}                &  1354.60      &  1170.16      &   865.15     & 710.17        & 880.80              &        697.45                       &  946.39    \\ &
\quad \cellcolor{blue!10}\textbf{w/ \method} &     \cellcolor{blue!10}{\textbf{1576.64}}   &   \cellcolor{blue!10}{\textbf{1387.50}}    &    \cellcolor{blue!10}{\textbf{1183.87}}    &      \cellcolor{blue!10}{\textbf{1146.48}}   &    \cellcolor{blue!10}{\textbf{1163.38}}           &   \cellcolor{blue!10}{\textbf{922.96}}                            &  \cellcolor{blue!10}{\textbf{1230.14}}  \\ & 
\quad\quad $\Delta$ & \color{darkgreen}{+222.04}  & \color{darkgreen}{+105.02}  & \color{darkgreen}{+296.10} & \color{darkgreen}{+434.15} & \color{darkgreen}{+250.79} & \color{darkgreen}{+173.12} & \color{darkgreen}{+250.00} \\
\bottomrule    
\end{tabular}
\label{tab:res_len}
\end{table*}

\subsubsection{Performance in 1.5B Base Model}
We first analyze the experimental results under the 1.5B base model, as shown in Table~\ref{tab:main_res_pass} and~\ref{tab:main_res_avg}. We observe that \method achieves the best results in $17$ out of $18$ tasks. Notably, \method obtains the best average result of $68.57$ across all tasks under the upper bound metric of \textit{Pass@K}, outperforming the second-best method (GRPO w/ Mask) by $3.26$ points.
Furthermore, for the average performance metric of \textit{Avg@K}, \method also achieves the best results in $17$ out of $18$ tasks, attaining an average result of $44.01$ across all $18$ tasks. These results demonstrate that \method not only improves the upper bound of reasoning performance but also achieves the best average performance.
Meanwhile, we observe that GRPO w/ high entropy reward fails to improve model performance, whereas constraint-based approaches such as GRPO w/ Mask and w/ Clip effectively enhance reasoning capability. While directly maximizing entropy (w/ En) shows some advantages on the \textit{Pass@K} metric, its average performance remains unsatisfactory, demonstrating its instability. In contrast, \method not only improves the upper bound of reasoning but also maintains stable average performance.

\subsubsection{Performance in Different Base Models}
To further validate the effectiveness of \method, we conduct additional experiments on Qwen2.5-Math-7B, a larger-scale base model with higher parameter capacity but without specialized reasoning function. As presented in Table~\ref{tab:res_larger_model}, \method demonstrates superior performance compared to all three baseline methods, achieving the best results across all $6$ benchmarks for the \textit{Pass@16} metric. For the \textit{Avg@16} metric, our method attains the best performance on $4$ benchmarks and second-best on Minerva benchmark. Notably, \method achieves substantial improvements over GRPO on \textit{Pass@16}, indicating that our approach effectively elevates the upper bound of the model's reasoning capability rather than merely optimizing within existing performance constraints. The consistent improvements observed across both the $1.5$B and $7$B base models provide strong evidence for the effectiveness of \method. These results demonstrate that our entropy-guided learning paradigm scales effectively to larger model architectures and successfully enhances reasoning capabilities across different model scales.

Moreover, we compare our method with GRPO on Qwen3-14B base model, with detailed results presented in Table~\ref{tab:res_larger_model_qwen3}. \method outperforms GRPO across all six benchmarks on both metrics, demonstrating the effectiveness of our approach on large-scale models. Notably, our method achieves a perfect score of $100$ on the MATH500 pass@16 metric, representing optimal performance on this benchmark. \textbf{Collectively, the consistent improvements across three different base models provide strong evidence that \method effectively enhances the reasoning capabilities of LLMs across different models.}

\begin{figure*}[t]
\centering
\includegraphics[width=0.24\linewidth]{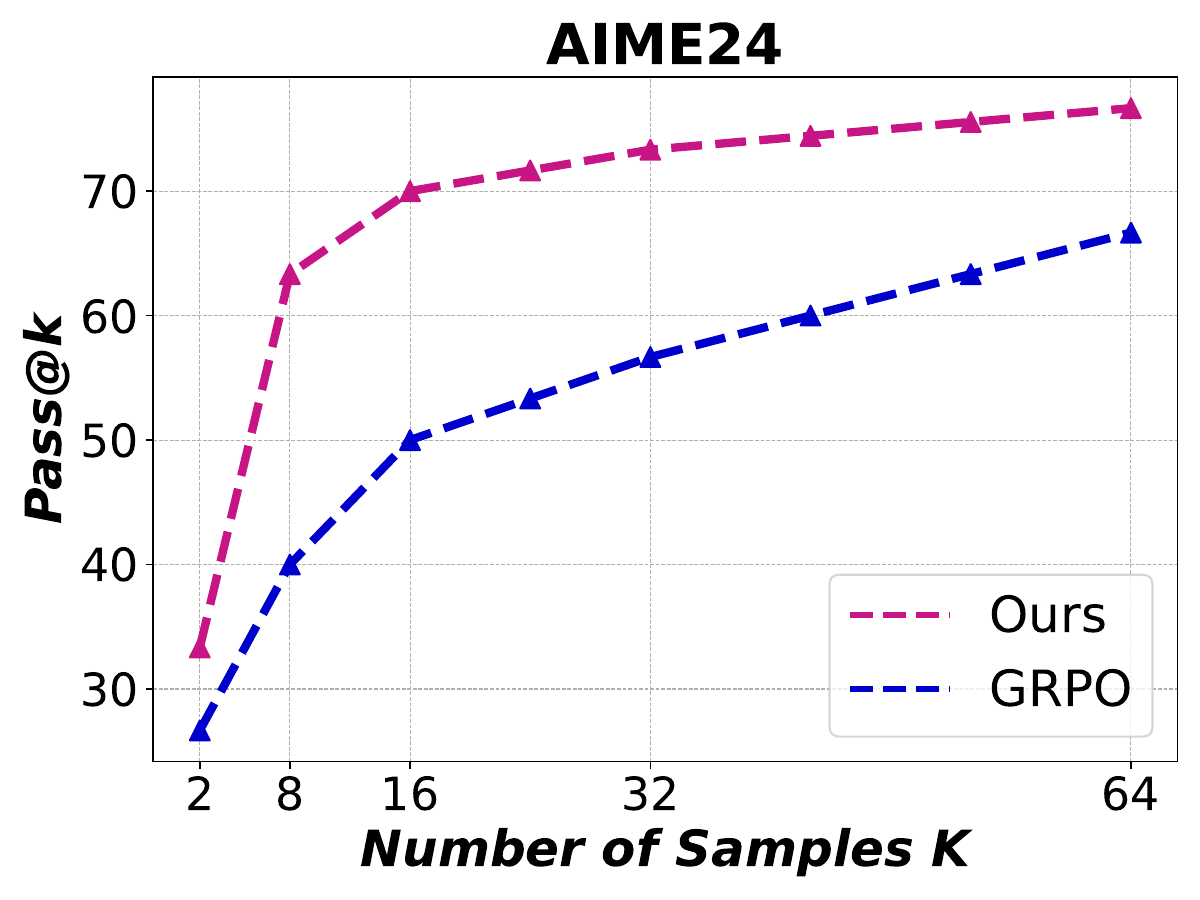}
\includegraphics[width=0.24\linewidth]{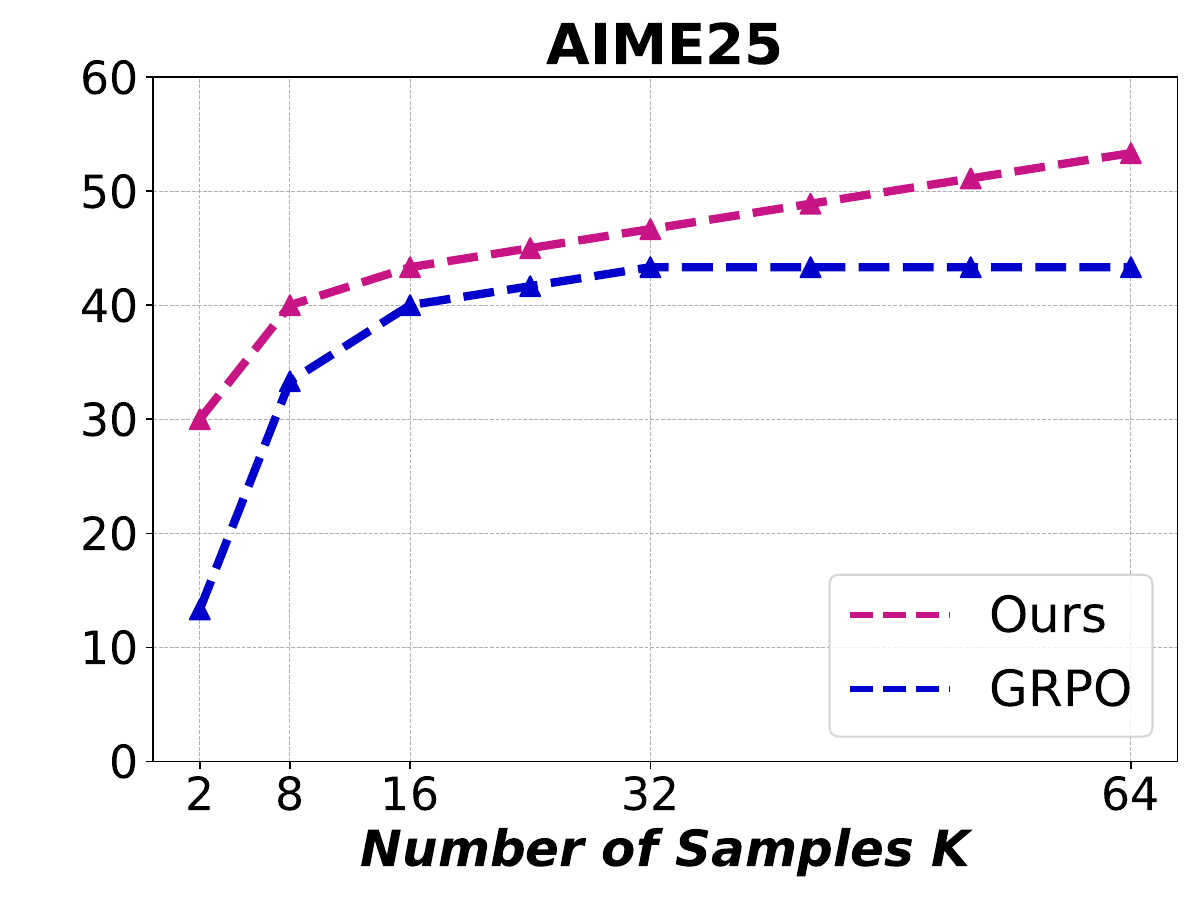}
\includegraphics[width=0.24\linewidth]{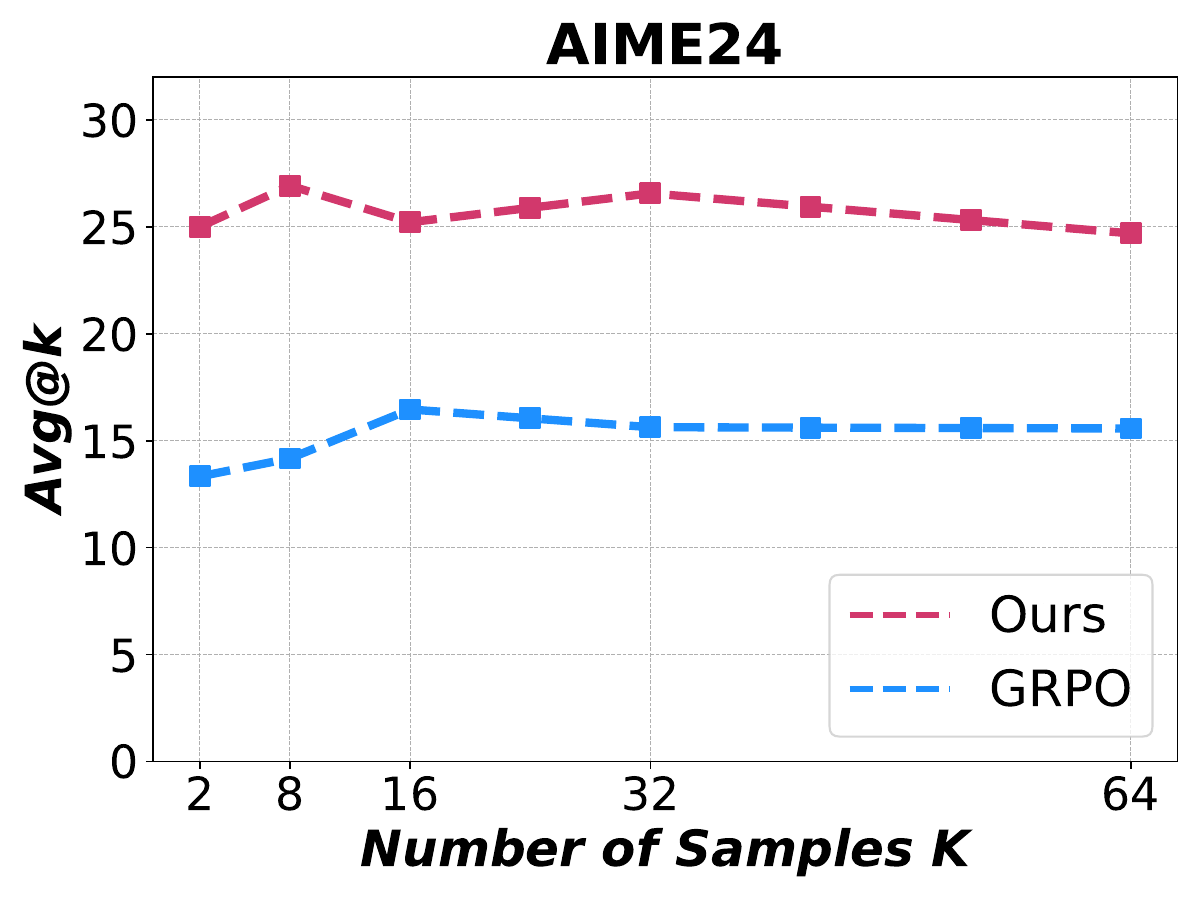}
\includegraphics[width=0.24\linewidth]{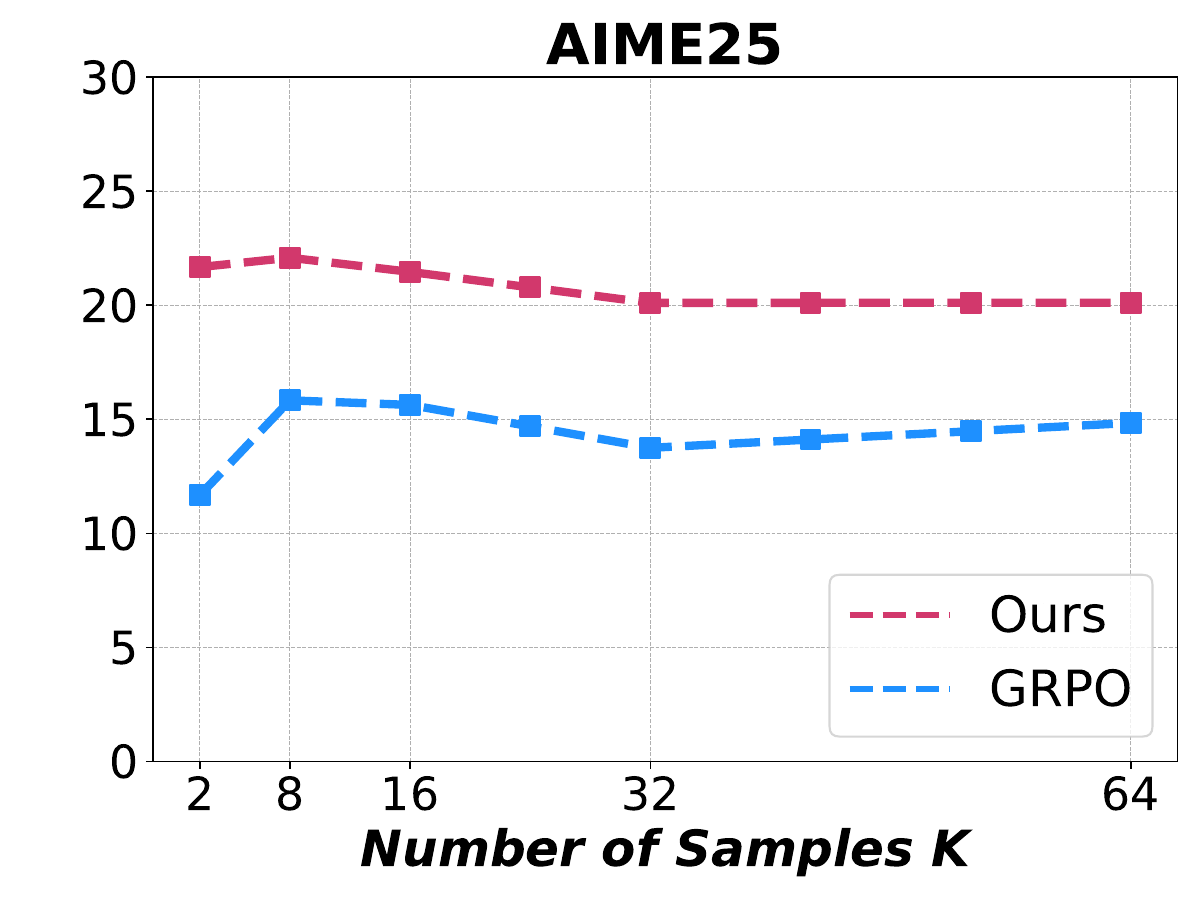}
\includegraphics[width=0.24\linewidth]{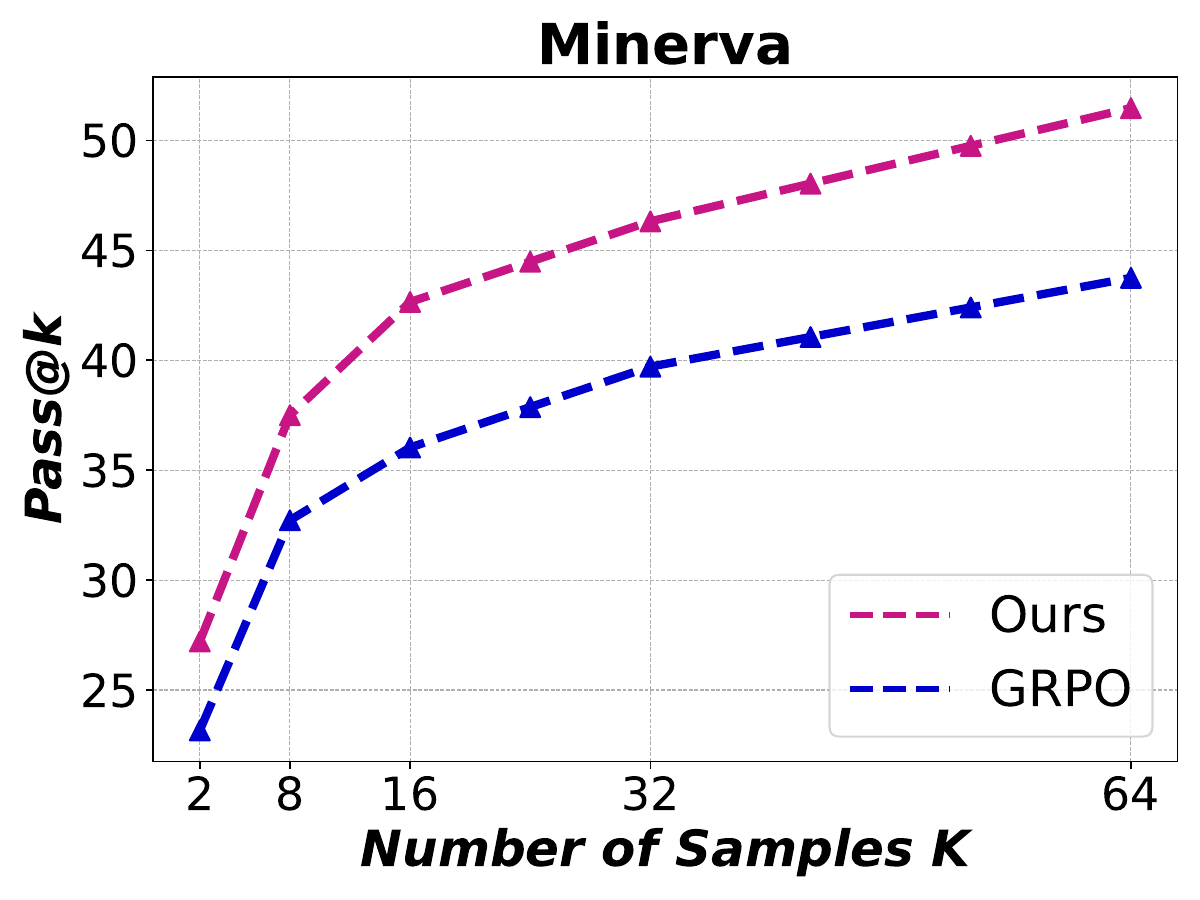}
\includegraphics[width=0.24\linewidth]{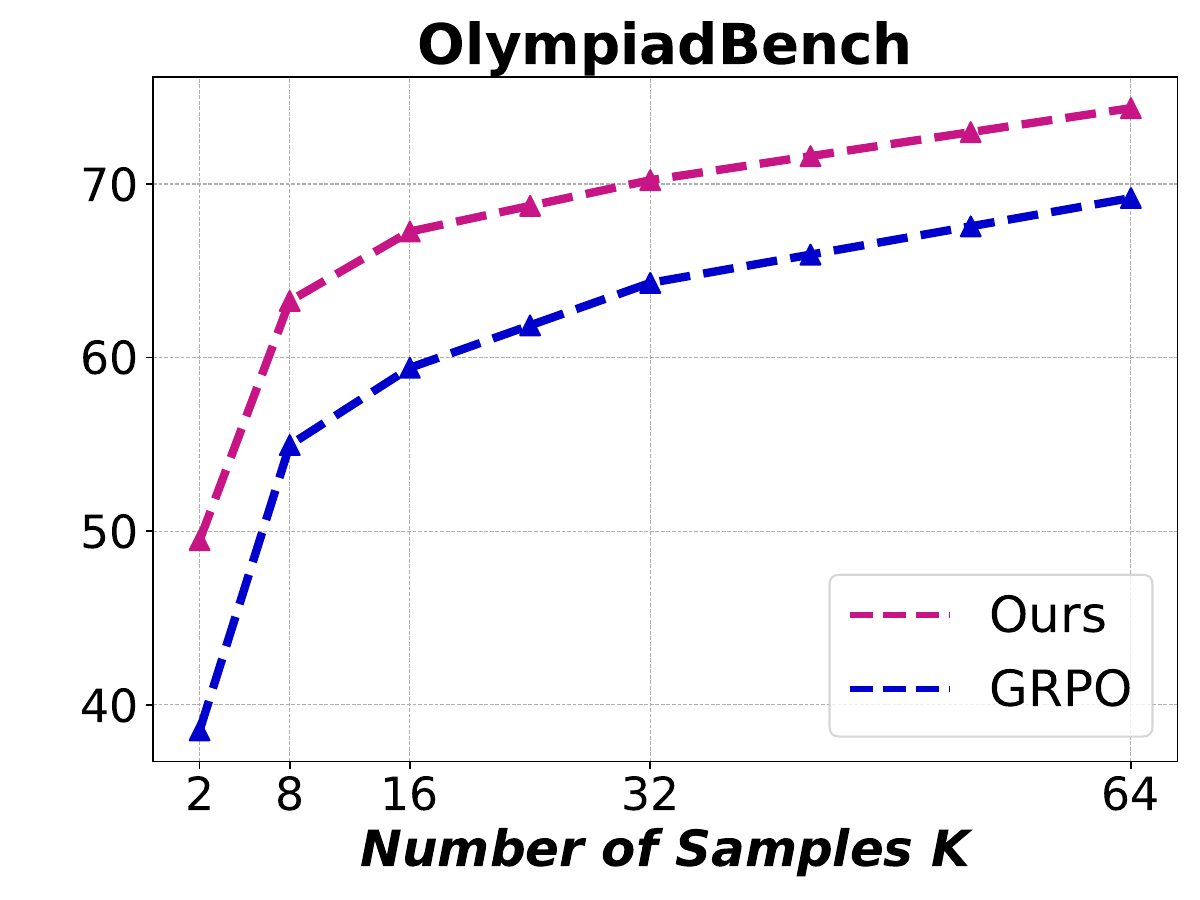}
\includegraphics[width=0.24\linewidth]{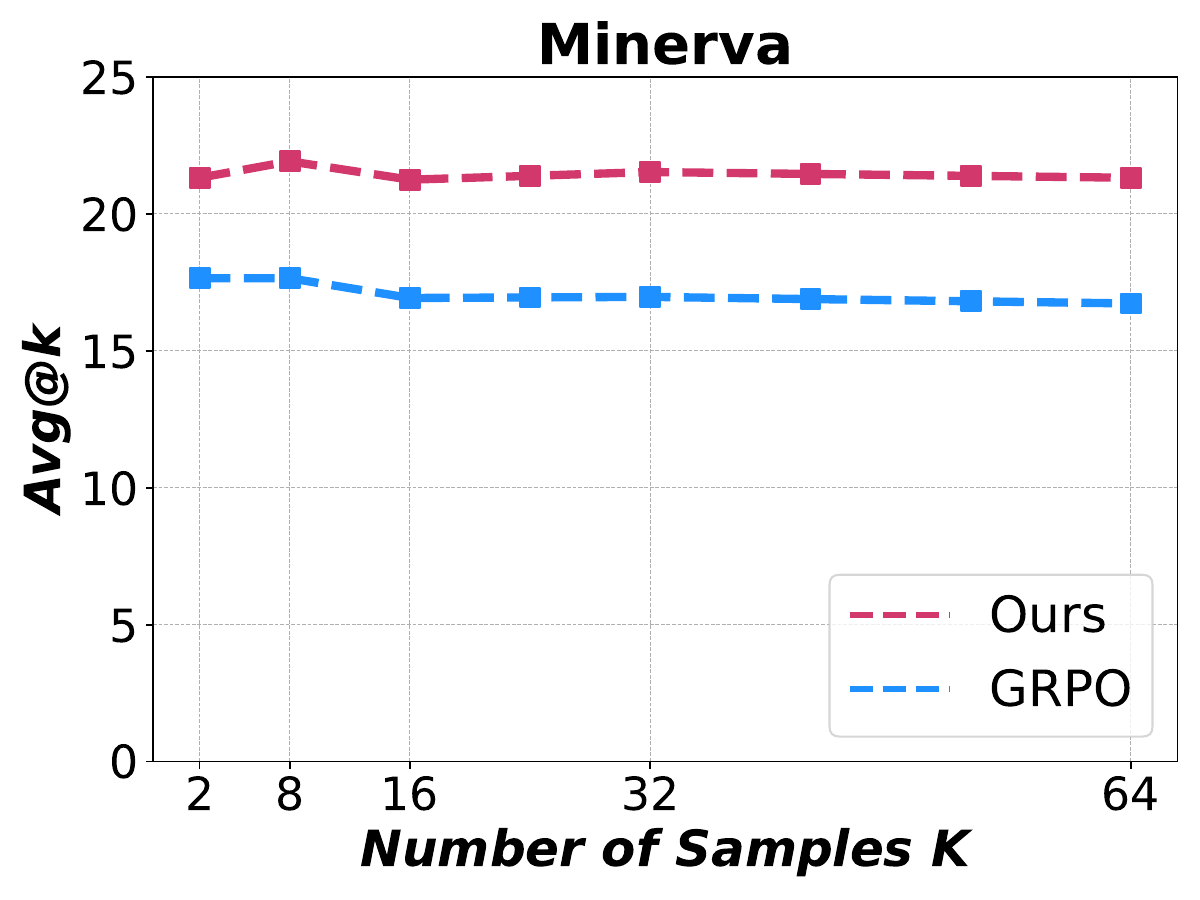}
\includegraphics[width=0.24\linewidth]{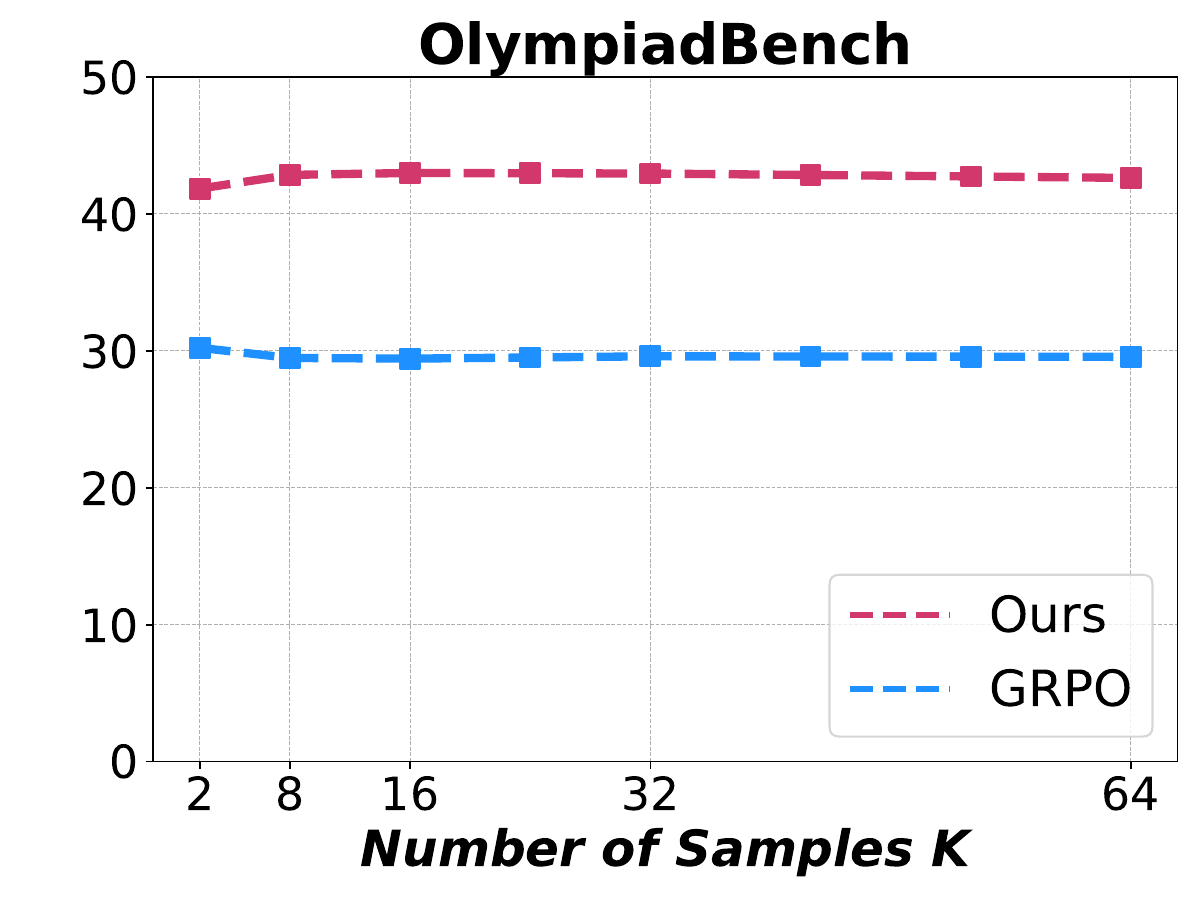}
\includegraphics[width=0.24\linewidth]{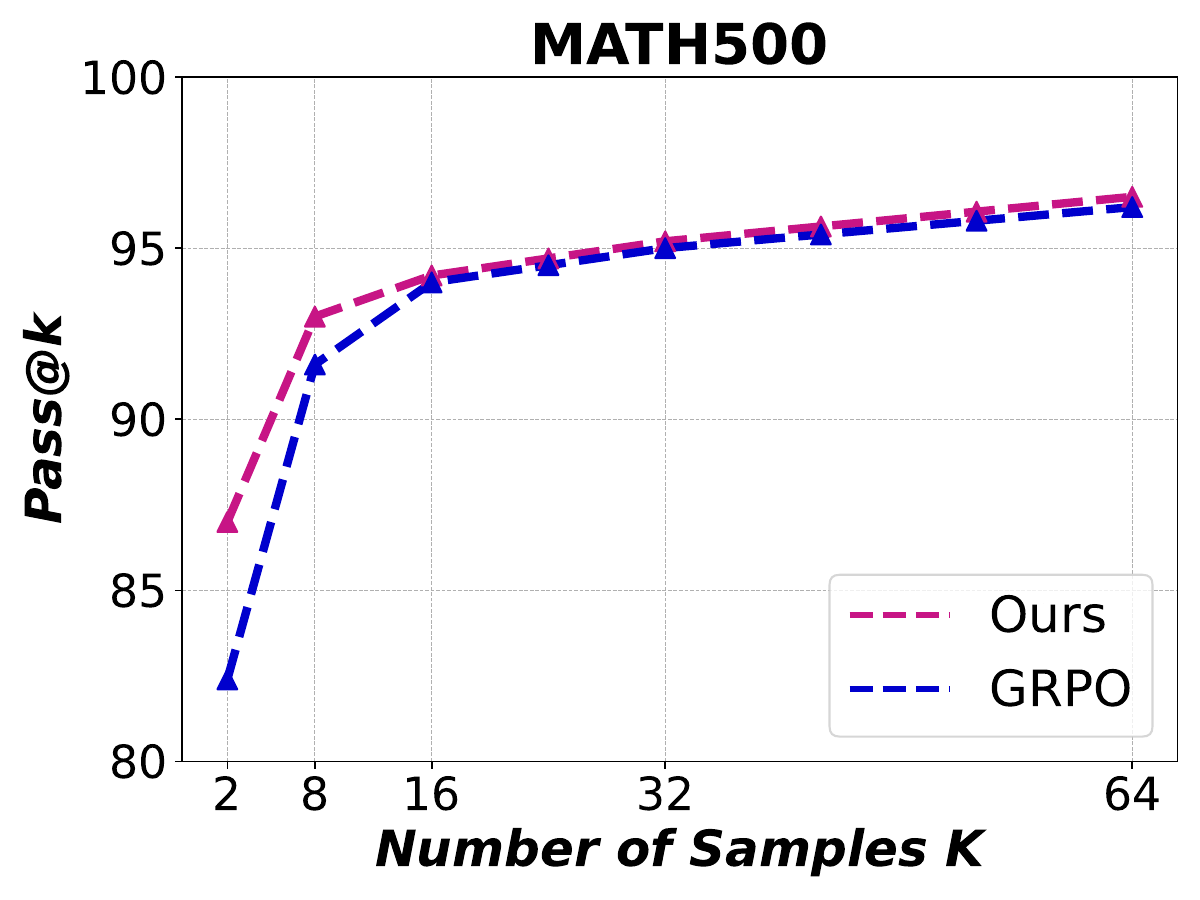}
\includegraphics[width=0.24\linewidth]{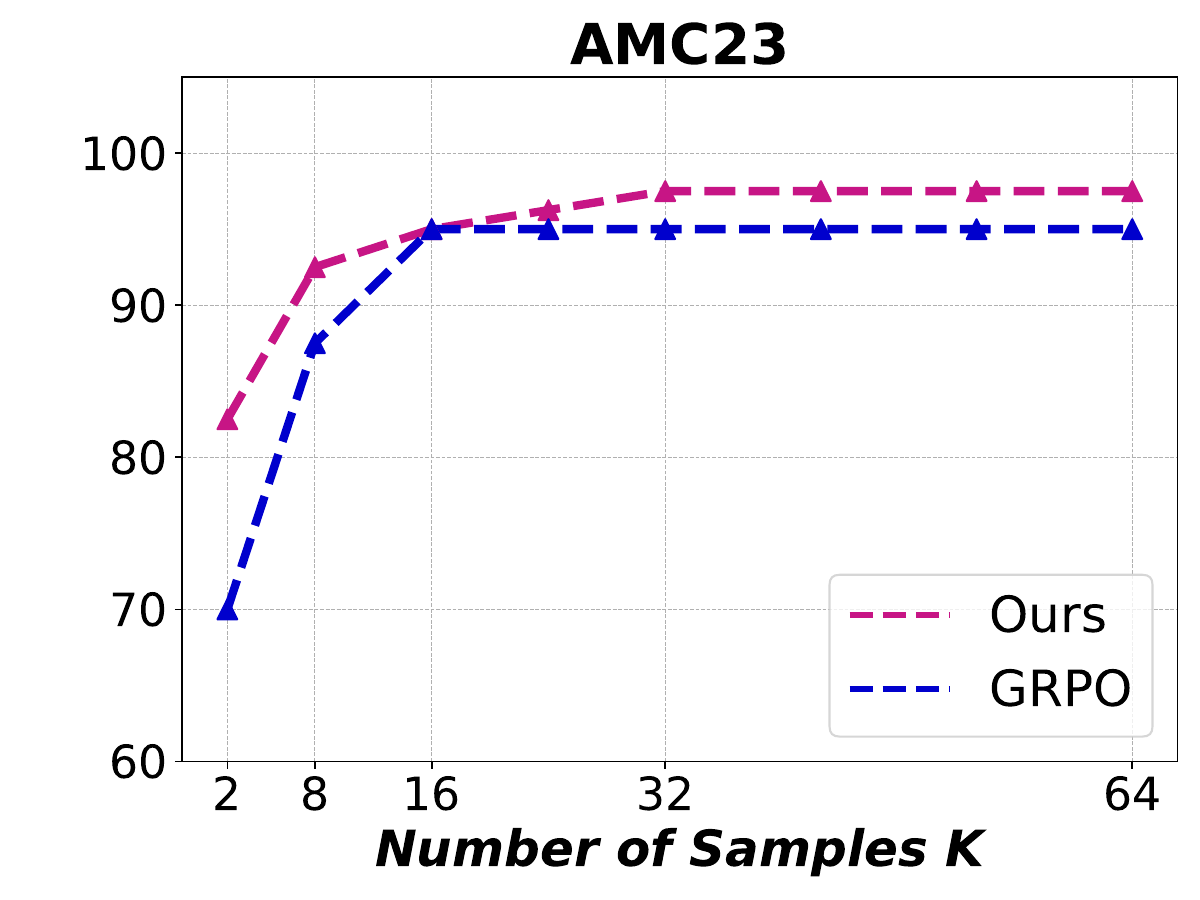}
\includegraphics[width=0.24\linewidth]{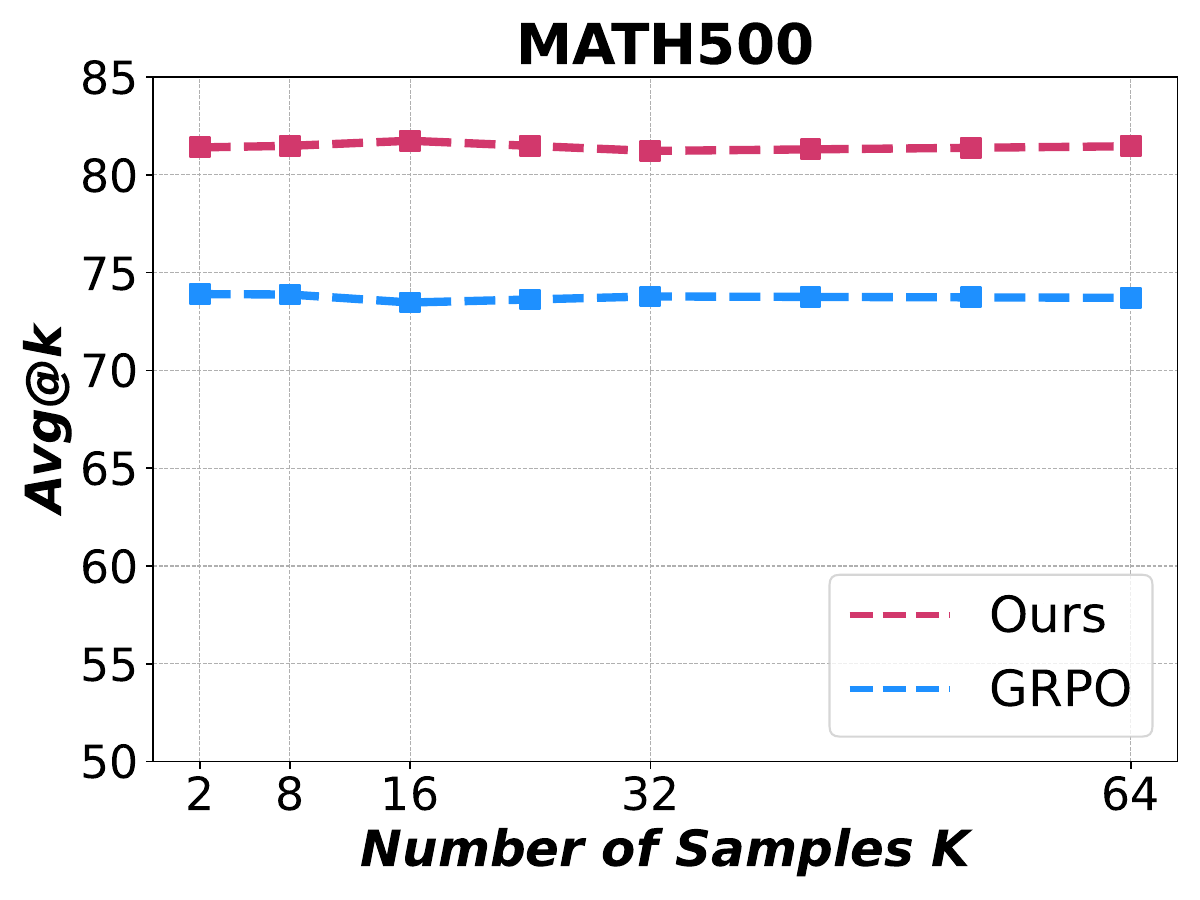}
\includegraphics[width=0.24\linewidth]{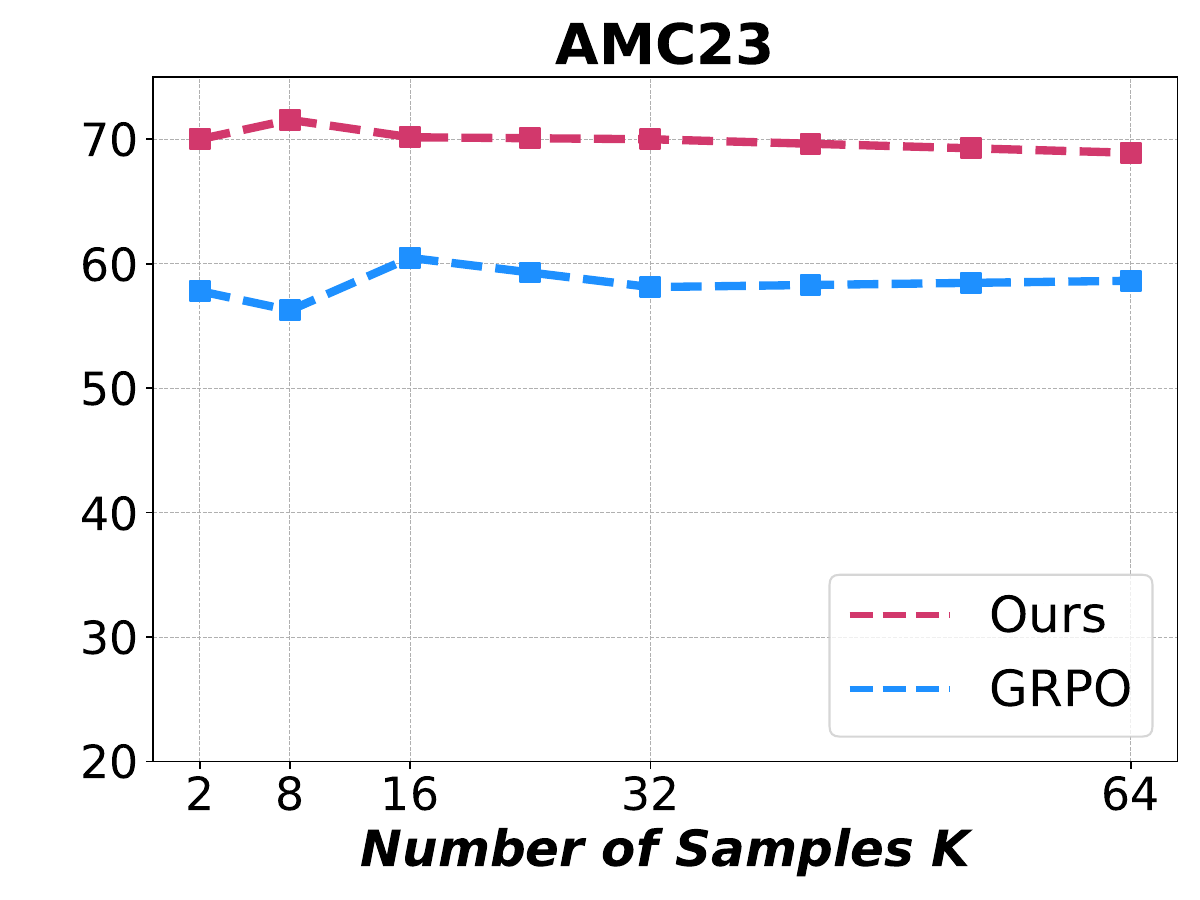}
\caption{\textit{Pass@K} and \textit{Avg@K} curves of \method compared with GRPO across $6$ benchmarks with the increase of number of samples $K$ from $1.5$B base models.}
\label{fig:pass@k_avg@k_curves}
\end{figure*}

\subsubsection{Performance in Response Length}
We further analyze the performance in response length as shown in Table~\ref{tab:res_len}. We find that \method achieves the best \textit{Len@16} results across all six benchmarks under the $1.5$B base model. Notably, \method generates responses that are $1003.85$ tokens longer on average compared to the second-best method.
For the non-reasoning-optimized Qwen2.5-Math-7B base model, \method demonstrates consistent advantages in eliciting longer reasoning chains compared to both baselines. Our method achieves an average response length of $1230.14$ tokens, surpassing GRPO and GRPO with entropy by $250.00$ and $283.75$ tokens respectively. The improvements are especially substantial on MATH500 ($+434.15$ tokens) and AMC23 ($+296.10$ tokens), indicating that \method effectively stimulates deeper reasoning even in models without specialized reasoning pre-training. These improvements in reasoning length provide strong evidence that \method successfully encourages more comprehensive exploration of solution spaces and multi-step reasoning processes. The correlation between increased response length and improved accuracy metrics validates that longer responses reflect genuine reasoning enhancement rather than verbosity.

Furthermore, we evaluate the scalability of \textit{Pass@k} performance with respect to increasing sample size $K$. As illustrated in Fig.~\ref{fig:pass@k_avg@k_curves}, we observe that \method sustains performance improvements even at large $K$ values, whereas most baselines, including GRPO, exhibit performance saturation. At smaller $K$ values, performance increases markedly as $K$ grows; conversely, at larger $K$ values, performance asymptotically approaches a stable plateau. Although the performance differential between \method and GRPO becomes negligible on the MATH500 benchmark at large 
$K$ values, these results demonstrates the enhanced capacity of \method to elevate the upper bound of reasoning performance compared to GRPO.
Furthermore, with respect to average performance of \textit{Avg@K}, \method exhibits robust and consistent performance across the entire range of $K$ values, uniformly surpassing GRPO irrespective of sample size configuration.

\begin{figure}[t]
\centering
\includegraphics[width=1\linewidth]{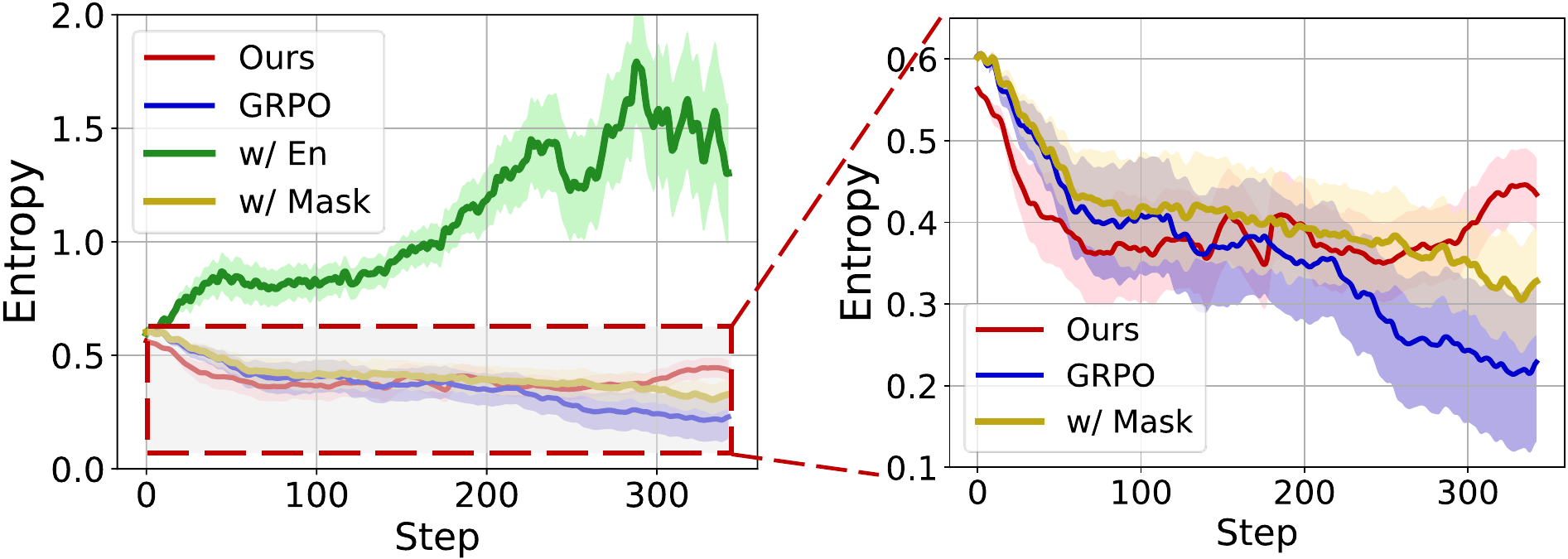}
\caption{The learning curves of entropy changes during learning process from $1.5$B models. w/ En means method of GRPO with entropy, and w/ Mask means GRPO with mask low entropy tokens. The shadow of line is the standard error.}
\label{fig:entropy}
\end{figure}

\subsection{Property Analysis}
\subsubsection{Entropy Changes}
As illustrated in Fig.~\ref{fig:entropy}, we observe that GRPO with entropy maintains persistently elevated entropy throughout training, culminating in entropy explosion and consequent training instability. Conversely, the remaining three methods demonstrate entropy reduction. Notably, GRPO exhibits precipitous entropy decline while GRPO w/ Mask demonstrates progressive entropy degradation, whereas \method achieves controlled entropy changes. During the first curriculum learning stage, entropy declines rapidly owing to the relative simplicity of the training tasks. In the subsequent stage, entropy increases effectively, thereby augmenting the model's exploratory capacity. These observations validate that our method successfully mitigates entropy collapse by maintaining entropy stability, avoiding both persistent elevation and continuous deterioration.

\subsubsection{Asymptotic Performance}
\begin{figure}[t]
\centering
\includegraphics[width=0.49\linewidth]{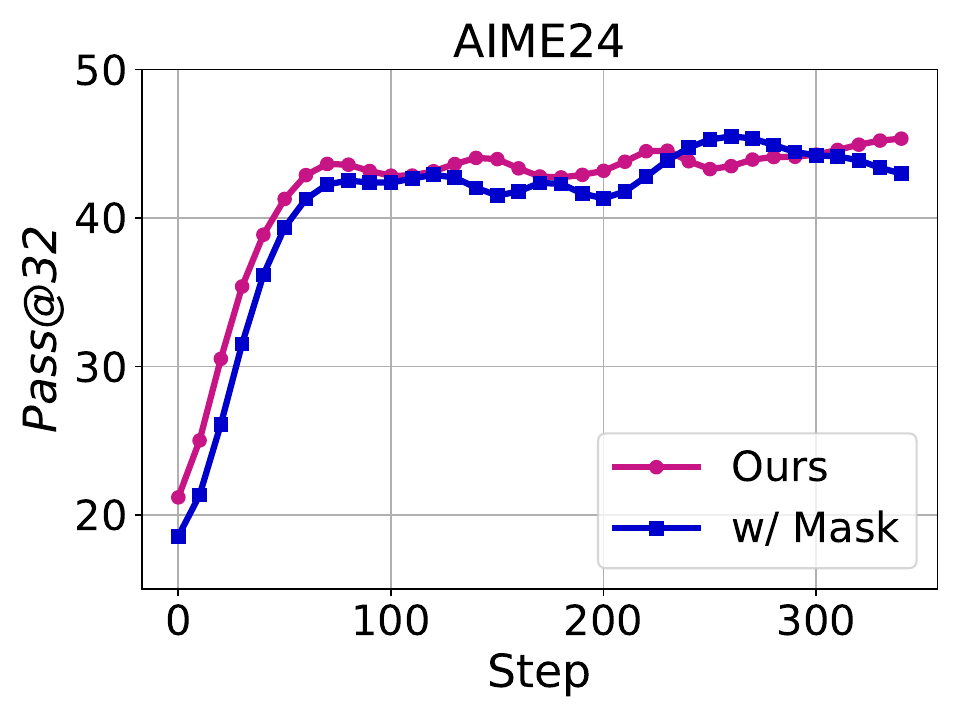}
\includegraphics[width=0.49\linewidth]{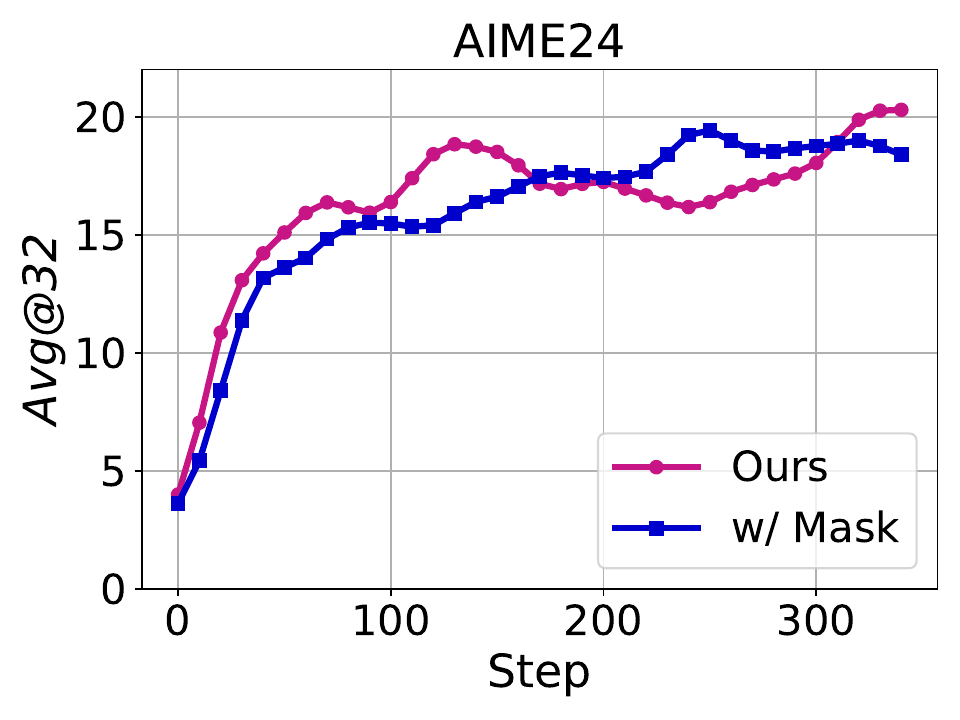}
\caption{The compared performance on "\textit{Avg@32}" and "\textit{Pass@32}" during training with DeepSeek-R1-Distill-Qwen-1.5B base model. w/ Mask means the method of GRPO with mask low entropy tokens.}
\label{fig:qwen_comp_1.5B}
\end{figure}

We analyze the asymptotic performance throughout the training process to examine the learning dynamics of \method. As illustrated in Fig.~\ref{fig:qwen_comp_1.5B}, we compare \method with the second-best baseline (GRPO w/ Mask) on the DeepSeek-R1-Distill-Qwen-1.5B model. The training exhibits two distinct phases of curriculum learning: In Stage 1 (steps 0-200), \method demonstrates superior learning efficiency, achieving faster performance improvements compared to the baseline. At Stage 2 (after step 200), \method experiences a transient performance decline due to the increased complexity of the training data introduced by our curriculum learning strategy. However, throughout the later phase of Stage 2, \method exhibits substantial performance recovery and ultimately surpasses the baseline method. This learning trajectory validates the effectiveness of our curriculum learning-driven token-level optimization approach, demonstrating that \method not only accelerates initial learning but also maintains robust improvement capacity when confronted with progressively challenging data.

\begin{figure}[t]
\centering
\includegraphics[width=0.49\linewidth]{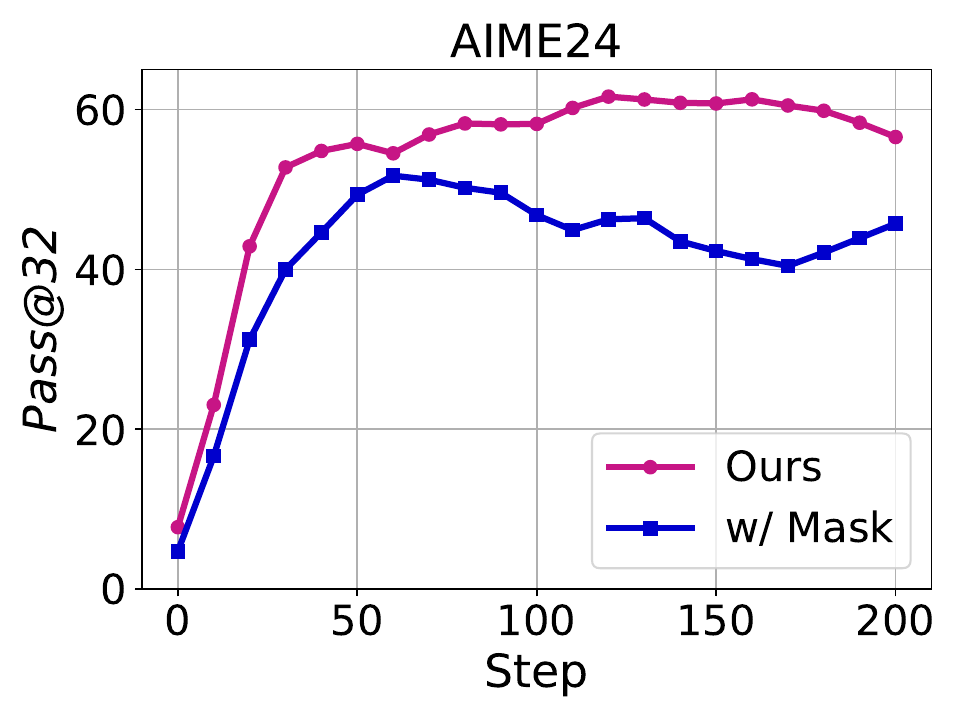}
\includegraphics[width=0.49\linewidth]{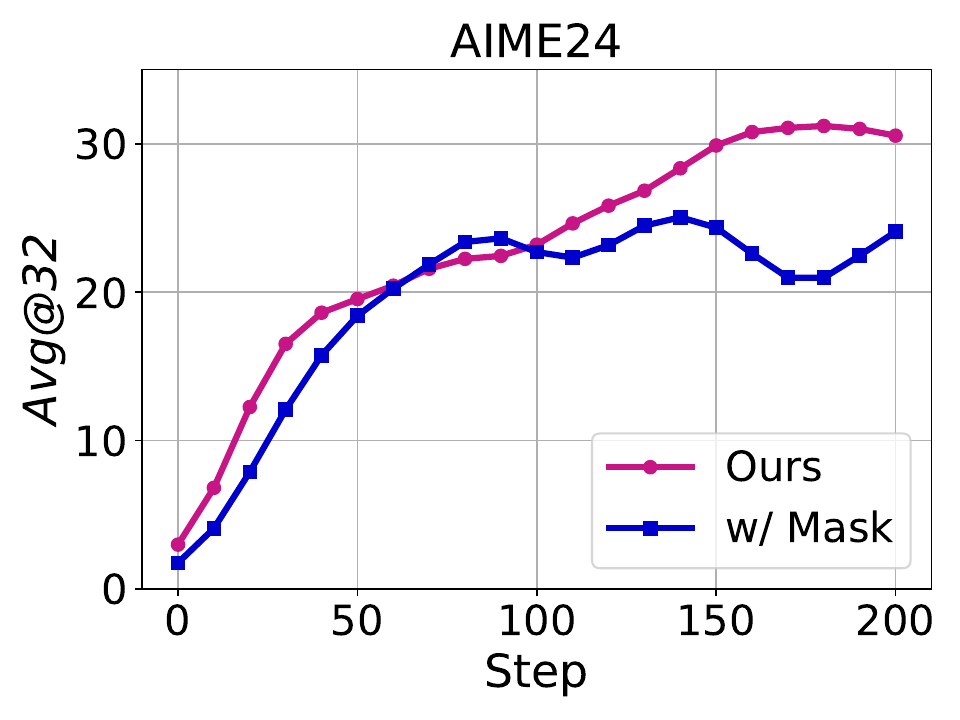}
\caption{The compared performance on "\textit{Avg@32}" and "\textit{Pass@32}" during training with Qwen2.5-Math-7B base model.}
\label{fig:qwen_comp_7B}
\end{figure}

For non-reasoning-optimized Qwen2.5-Math-7B base model, as shown in Fig.~\ref{fig:qwen_comp_7B}, \method achieves consistent and stable performance improvements throughout the entire training process on both \textit{Pass@32} and \textit{Avg@32} metrics compared to GRPO w/ Mask. Notably, unlike the $1.5$B model, the $7$B model exhibits monotonic performance growth without the transient decline observed in Stage 2, suggesting that larger model capacity provides greater resilience to curriculum difficulty increases. These results demonstrate that \method delivers stable and reliable performance enhancements across different model scales.

\subsubsection{Semantic Entropy Analysis}
\begin{figure}[t]
\centering
\includegraphics[width=1\linewidth]{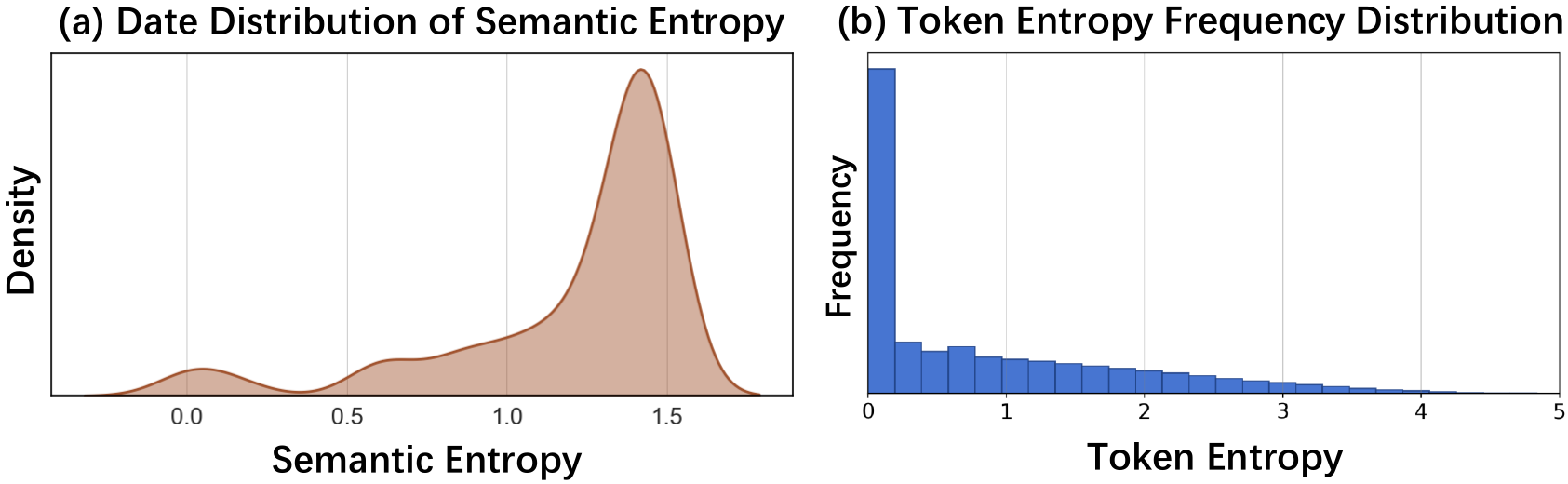}
\caption{The data distribution of semantic entropy for the training dataset and the token entropy frequency distribution.}
\label{fig:dis_entropy}
\end{figure}

In \method, we calculate the semantic entropy of the training data and design curriculum learning based on semantic entropy. As illustrated in Fig.~\ref{fig:dis_entropy}(a), we present the distribution of semantic entropy across the dataset. We observe that the majority of data exhibits semantic entropy concentrated between $1.0$ and $1.75$. Semantic entropy reflects the degree of uncertainty in model responses to identical questions, where higher semantic entropy indicates greater task difficulty. Based on this distribution, we organize the data in ascending order of entropy values for model training. The corpus is partitioned into two distinct subsets, thereby establishing a two-stage curriculum learning framework.

\begin{figure}[t]
\centering
\includegraphics[width=1\linewidth]{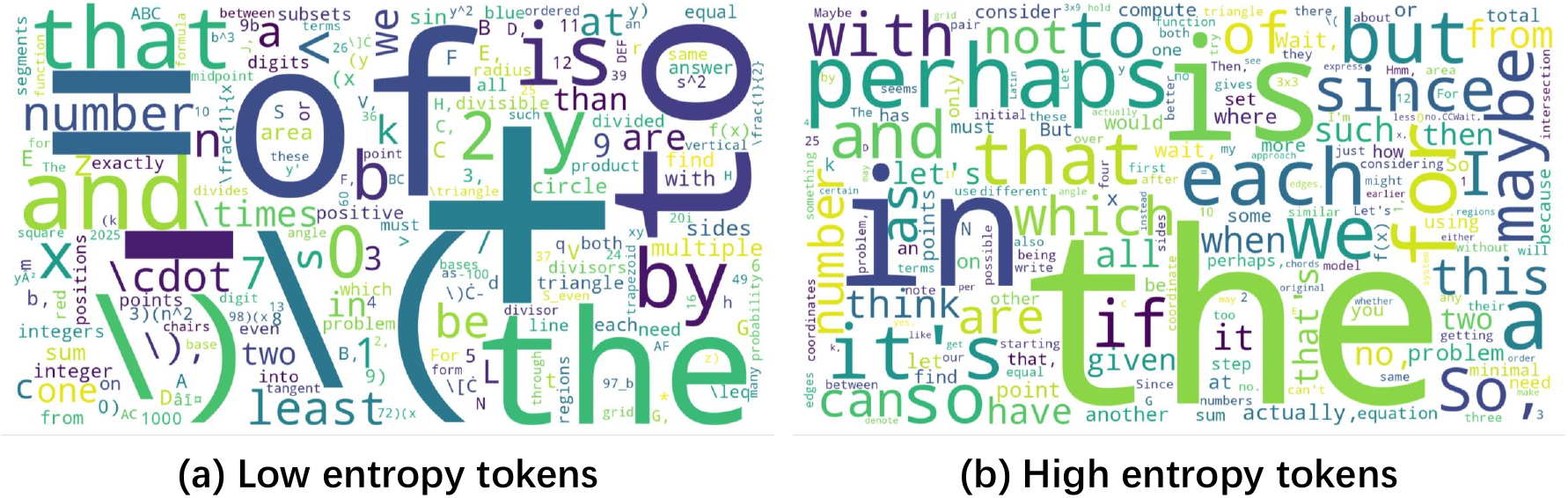}
\caption{Words Cloud for low entropy tokens and high entropy tokens.}
\label{fig:word_cloud}
\end{figure}

\subsubsection{Token-Level Entropy Analysis}
We conduct a fine-grained analysis of entropy distribution at the token-level. As illustrated in Fig.~\ref{fig:dis_entropy}(b), a substantial proportion of tokens are concentrated in the low-entropy region, indicating high prediction confidence for these tokens. Furthermore, as depicted in the word cloud visualization in Fig.~\ref{fig:word_cloud}, we observe that low-entropy tokens predominantly consist of common words such as "the," "of," and "to," which contribute minimally to reasoning enhancement. Notably, high-entropy tokens also contain a mixture of such function words alongside more semantically meaningful tokens. This observation reveals a critical insight: \textbf{naively encouraging exploration by targeting all high-entropy tokens risks diluting the learning signal with linguistically trivial tokens, thereby undermining effective policy exploration.} In contrast, our method strategically focuses on addressing problematic low-entropy tokens those that prematurely converge and constrain reasoning diversity, while avoiding interference from semantically vacuous tokens. This targeted approach ensures that entropy regularization enhances exploratory behavior in reasoning-critical tokens rather than amplifying noise from linguistic artifacts.

\subsubsection{Curriculum Design Analysis}
\label{sec:curr_analysis}

We further analyze the performance of \method under different curriculum designs, with detailed results presented in Table~\ref{tab:cur}. Compared to no curriculum and three-stage curriculum, the two-stage curriculum learning achieves the best performance across all six benchmarks. Notably, two-stage curriculum attains the highest average Avg@16 scores across all benchmarks, demonstrating more stable and consistent learning under this configuration. These results suggest that a moderate curriculum progression provides the optimal balance between gradual difficulty adaptation and training efficiency, allowing the model to build reasoning capabilities without prolonging the learning process.

\begin{table*}[t]
 \caption{Curriculum Design Analysis on six benchmarks under \textit{Pass@16} and \textit{Avg@16} using $1.5$B base models. We bold the best results.} 
\renewcommand{\arraystretch}{1.2}
\setlength{\tabcolsep}{3.2pt} 
    \centering 
\begin{tabular}{lcccccccccccc}
\toprule
\textbf{}                          & \multicolumn{2}{c}{\textbf{AIME24}} & \multicolumn{2}{c}{\textbf{AIME25}} & \multicolumn{2}{c}{\textbf{MATH500}}  & \multicolumn{2}{c}{\textbf{Minerva}}  & \multicolumn{2}{c}{\textbf{AMC23}} &  \multicolumn{2}{c}{\textbf{OlympiadBench}}\\ \cmidrule(lr){2-3} \cmidrule(lr){4-5} \cmidrule(lr){6-7} \cmidrule(lr){8-9} \cmidrule(lr){10-11} \cmidrule(lr){12-13}
& \textit{Pass@16}  & \textit{Avg@16} & \textit{Pass@16}  & \textit{Avg@16} & \textit{Pass@16} & \textit{Avg@16} & \textit{Pass@16}  & \textit{Avg@16}  & \textit{Pass@16}     & \textit{Avg@16}     & \textit{Pass@16}  & \textit{Avg@16}  \\ \bottomrule
\quad \textbf{No Curriculum} &  \textbf{63.33} &  25.00 &  40.00 & 20.42 &  93.80 &  81.60 &  \textbf{43.38} &  21.07 &  \textbf{95.00} &  69.22 & 66.22 & 42.72  \\ 
\rowcolor{blue!10} 
\quad \textbf{Two-Stage}  & \textbf{63.33}  & \textbf{25.21} & \textbf{43.33} & \textbf{21.46} & \textbf{94.20} & \textbf{81.74} & 42.65 & \textbf{21.25} & \textbf{95.00} & \textbf{70.16} & \textbf{67.26} & \textbf{42.99} 
\\ 
\quad \textbf{Three-Stage}   & 60.00   &  23.54   & \textbf{43.33}  & 18.75  & \textbf{94.20}  & 80.75  & 39.71  & 20.73  & \textbf{95.00}  & 67.81  & 66.37  & 41.94                     \\ 
\bottomrule
\end{tabular}
\label{tab:cur}
\end{table*}

\begin{table*}[t]
 \caption{Hyperparameter analysis on six benchmarks under \textit{Pass@16} and \textit{Avg@16} using $1.5$B base models. We bold the best results and underline the sub-optimal results.} 
\renewcommand{\arraystretch}{1.2}
\setlength{\tabcolsep}{3.2pt} 
    \centering 
\begin{tabular}{lcccccccccccc}
\toprule
\textbf{}                          & \multicolumn{2}{c}{\textbf{AIME24}} & \multicolumn{2}{c}{\textbf{AIME25}} & \multicolumn{2}{c}{\textbf{MATH500}}  & \multicolumn{2}{c}{\textbf{Minerva}}  & \multicolumn{2}{c}{\textbf{AMC23}} &  \multicolumn{2}{c}{\textbf{OlympiadBench}}\\ \cmidrule(lr){2-3} \cmidrule(lr){4-5} \cmidrule(lr){6-7} \cmidrule(lr){8-9} \cmidrule(lr){10-11} \cmidrule(lr){12-13}
& \textit{Pass@16}  & \textit{Avg@16} & \textit{Pass@16}  & \textit{Avg@16} & \textit{Pass@16} & \textit{Avg@16} & \textit{Pass@16}  & \textit{Avg@16}  & \textit{Pass@16}     & \textit{Avg@16}     & \textit{Pass@16}  & \textit{Avg@16}  \\ \bottomrule
\quad \textbf{en: 70\%, cov: 0.0002} &  \textbf{73.33} &  24.17 &  40.00 & 17.92 &  94.00 &  81.10 &  41.54 &  20.40 &  95.00 &  68.13 & 65.93 & 42.52  \\ 
\quad \textbf{en: 90\%, cov: 0.0002}  & 66.67  & 24.58 & 40.00 & 18.75 & 93.80 & 80.50 & \textbf{43.01} & \textbf{21.83} & 95.00 & 68.75 & 66.22 & 42.65 
\\ 
\quad \textbf{en: 80\%, cov: 0.0001}   & 56.67   &  21.88   & 40.00  & 21.25  & \textbf{95.60}  & 81.60  & 41.54  & 21.53  & 95.00  & 69.84  & 65.19  & 42.80                     \\ 
\quad \textbf{en: 80\%, cov: 0.002} & 56.67 & 24.38 & 40.00 & 18.33 & 93.20 & 81.55 & 41.54 & 21.14 & 95.00 & 68.44 & 66.07 & 42.25  \\ 
 \rowcolor{blue!10} 
\quad \textbf{en: 80\%, cov=0.0002}   & \underline{70.00}    & \textbf{25.21}  & \textbf{43.33}    & \textbf{21.46}  & \underline{94.20}   & \textbf{81.74}  & \underline{42.65}    & \underline{21.25}   & 95.00       & \textbf{70.16}      & \textbf{67.26}       & \textbf{42.99}  \\
\bottomrule
\end{tabular}
\label{tab:hps}
\end{table*}

\begin{table}[t]
 \caption{Generalization study on LiveCodeBench benchmark under $1.5$B base model. w/ means with. We compare \method with GRPO.} 
\renewcommand{\arraystretch}{1.2}
\setlength{\tabcolsep}{6pt} 
\centering
\begin{tabular}{lccccc}
\toprule
                          & \multicolumn{5}{c}{\textbf{{LiveCodeBench}}}                                                     \\ \cmidrule(lr){2-6} 
\textbf{Method}                    & \multicolumn{1}{c}{\textit{Pass@8}} & \multicolumn{1}{c}{\textit{Pass@16}} & \multicolumn{1}{c}{\textit{Avg@8}} & \multicolumn{1}{c}{\textit{Avg@16}} & \textbf{Avg.} \\ \bottomrule
\textbf{GRPO}                      & 40.11    &  44.20       &            22.51       &   22.59                         &        32.35                    \\
\rowcolor{blue!10} \quad \textbf{w/ \method} &   \textbf{42.84}        &   \textbf{48.07}         &    \textbf{24.56}      &     \textbf{24.21}                       &          \textbf{34.92}                  \\ 
\quad\quad $\Delta$ & \color{darkgreen}{+2.73} &\color{darkgreen}{+3.87} & \color{darkgreen}{+2.05} & \color{darkgreen}{+1.62} & \color{darkgreen}{+2.57}
\\
\bottomrule
\end{tabular}
\label{tab:generalizaion}
\end{table}

\begin{table*}[h]
 \caption{Ablation study on six benchmarks of \textit{Pass@16} and \textit{Avg@16} under $1.5$B base models. w/ means with. We bold the best results and underline the sub-optimal results.} 
\renewcommand{\arraystretch}{1.2}
\setlength{\tabcolsep}{3.8pt} 
    \centering 
\begin{tabular}{lcccccccccccc}
\toprule
\textbf{}                          & \multicolumn{2}{c}{\textbf{AIME24}} & \multicolumn{2}{c}{\textbf{AIME25}} & \multicolumn{2}{c}{\textbf{AMC23}} & \multicolumn{2}{c}{\textbf{MATH500}} & \multicolumn{2}{c}{\textbf{OlympiadBench}} & \multicolumn{2}{c}{\textbf{Minerva}} \\ \cmidrule(lr){2-3} \cmidrule(lr){4-5} \cmidrule(lr){6-7} \cmidrule(lr){8-9} \cmidrule(lr){10-11} \cmidrule(lr){12-13}
\textbf{Method}                    & \textit{Pass@16}  & \textit{Avg@16} & \textit{Pass@16}  & \textit{Avg@16} & \textit{Pass@16} & \textit{Avg@16} & \textit{Pass@16}  & \textit{Avg@16}  & \textit{Pass@16}     & \textit{Avg@16}     & \textit{Pass@16}  & \textit{Avg@16}  \\ \bottomrule
 \textbf{GRPO}                      & 50.00             & 16.46           & \underline{40.00}       & 15.63           & 95.00   & 60.47           & 94.00             & 73.47            & 59.41                & 29.43               & 36.03             & 16.93            \\
\quad \textbf{w/ Cur}                    & 56.67             & 20.42           & \underline{40.00}       & 19.17           & 95.00   & 68.91           & 93.00             & 81.15            & 65.63                & 42.65               & 39.34             & 20.38            \\
\quad \textbf{w/ Low}                    & 60.00             & 23.12           & \textbf{43.33}    & 17.71           & 95.00   & \underline{69.84}     & \textbf{94.20}    & 80.66            & \textbf{67.26}       & 41.29               & 42.28             & \underline{21.16}      \\
\quad \textbf{w/ Low\_Cov}               & \underline{63.33}       & \underline{25.00}     & \underline{40.00}       & \underline{20.42}     & 95.00   & 69.22           & \underline{93.80}       & \underline{81.60}      & \underline{66.22}          & \underline{42.72}         & \textbf{43.38}    & 21.07            \\ \rowcolor{blue!10}
\quad \textbf{w/ \method} & \textbf{70.00}    & \textbf{25.21}  & \textbf{43.33}    & \textbf{21.46}  & 95.00   & \textbf{70.16}  & \textbf{94.20}    & \textbf{81.74}   & \textbf{67.26}       & \textbf{42.99}      & \underline{42.65}       & \textbf{21.25}   \\ \bottomrule 
\end{tabular}
\label{tab:res_ablation}
\end{table*}

\subsubsection{Hyperparameter Analysis}
\label{sec:hps}
As presented in Table~\ref{tab:hps}, we conduct a systematic hyperparameter sensitivity analysis to identify the optimal configuration for \method. We investigate the impact of two key hyperparameters: the low-entropy token processing ratio (en) and the high-covariance processing ratio (cov). The results demonstrate that the configuration with en=$80\%$ and cov=$0.0002$ achieves superior overall performance, securing the best results on $3$ benchmarks (AIME25, MATH500, OlympiadBench) and second-best on $3$ benchmarks (AIME24, Minerva, AMC23) for \textit{Pass@16}, while attaining the best performance on $5$ out of $6$ benchmarks for \textit{Avg@16}. Consequently, we adopt this configuration as our default hyperparameter setting.

Moreover, we observe that \method maintains robust performance across various hyperparameter configurations, with performance variations remaining within a narrow range. For instance, \textit{Avg@16} scores on OlympiadBench fluctuate only between $42.25$ and $42.99$ across all configurations, while \textit{Pass@16} on AMC23 remains stable at $95\%$. This stability indicates that our method exhibits low hyperparameter sensitivity and demonstrates consistent generalization capability, rather than being critically dependent on precise hyperparameter tuning. Such robustness enhances the practical applicability of \method, as it reduces the computational overhead associated with extensive hyperparameter search in real-world deployment scenarios.

\begin{figure}
\centering
\includegraphics[width=0.49\linewidth]{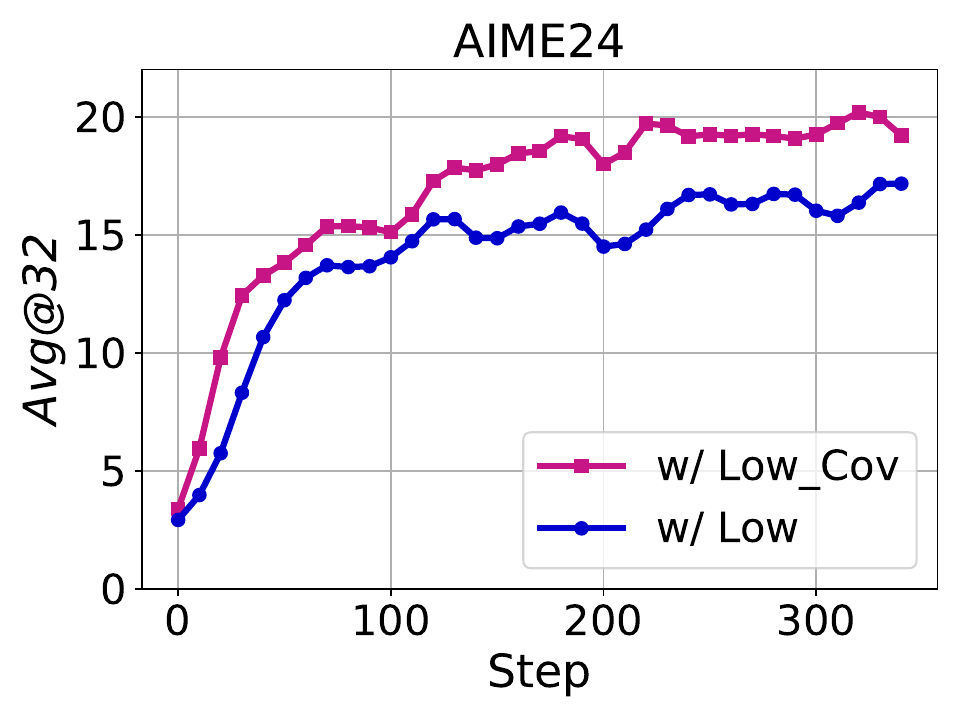}
\includegraphics[width=0.49\linewidth]{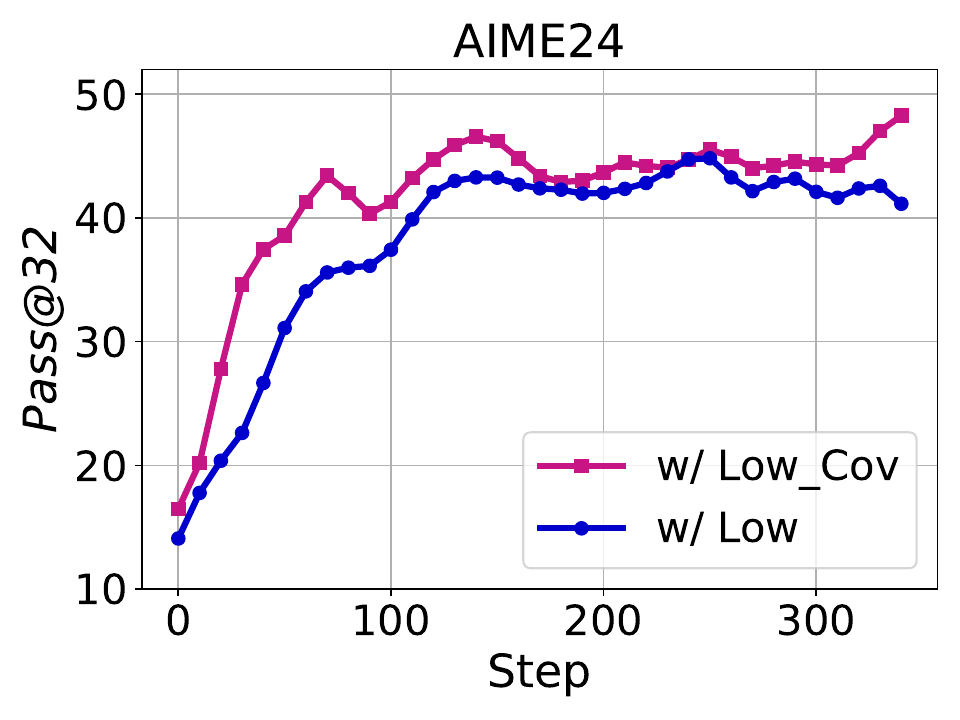}
\caption{The compared performance on "\textit{Avg@32}" and "\textit{Pass@32}" during training. w/ Low means the method of low entropy constraint. w/ Low\_Cov  means the method of high covariance constraint. }
\label{fig:abl_curve}
\end{figure}

\subsubsection{Generalization Ability}
To assess the cross-domain generalization capability of \method, we evaluate its performance on the LiveCodeBench benchmark~\cite{jain2025livecodebench}, a code generation testbed that differs substantially from the mathematical reasoning tasks employed during training. This evaluation enables us to examine whether the reasoning enhancements induced by \method transfer effectively to out-of-distribution problem domains. As presented in Table~\ref{tab:generalizaion}, \method consistently outperforms GRPO across all three evaluation metrics, providing empirical evidence that our approach cultivates generalizable reasoning capabilities that transcend the training task distribution. \textbf{These results suggest that \method's entropy-regularized learning paradigm facilitates the development of more robust and transferable reasoning strategies, rather than merely overfitting to domain-specific patterns in the training data.}

\subsection{Ablation Study}
We conduct an ablation study to analyze the effect of different components in \method, as shown in Table~\ref{tab:res_ablation}. The comparative analysis reveals that integrating curriculum learning yields substantial performance gains on the \textit{Avg@16} metric relative to GRPO, corroborating the progressive enhancement capacity afforded by curriculum learning. Analogously, the introduction of low-entropy token constraints and high-covariance constraints independently demonstrate performance improvements over the GRPO baseline. These enhancements are particularly salient on AIME24, OlympiadBench, and Minerva benchmarks. Furthermore, these methods achieve marked improvements on \textit{Pass@16}, substantiating that our proposed constraint-based optimization framework elevates the upper bound of reasoning capability.

Moreover, as illustrated in Fig.~\ref{fig:abl_curve}, we observe that jointly constraining high-covariance tokens with low-entropy tokens yields superior learning dynamics during training compared to constraining low-entropy tokens in isolation. This validates the effectiveness of our \textit{fine-grained} optimization strategy.

Notably, the integration of curriculum learning with our proposed optimization algorithm yields additional performance gains, particularly evident in average performance metrics. \method achieves superior performance by synergistically combining semantic entropy-based curriculum learning at the data level with token-level entropy optimization at the algorithmic level. This dual-perspective approach that organizes training data according to semantic entropy distribution while constraining token-level entropy during optimization, effectively enhances reasoning ability.

\begin{figure*}
\centering
\includegraphics[width=1\linewidth]{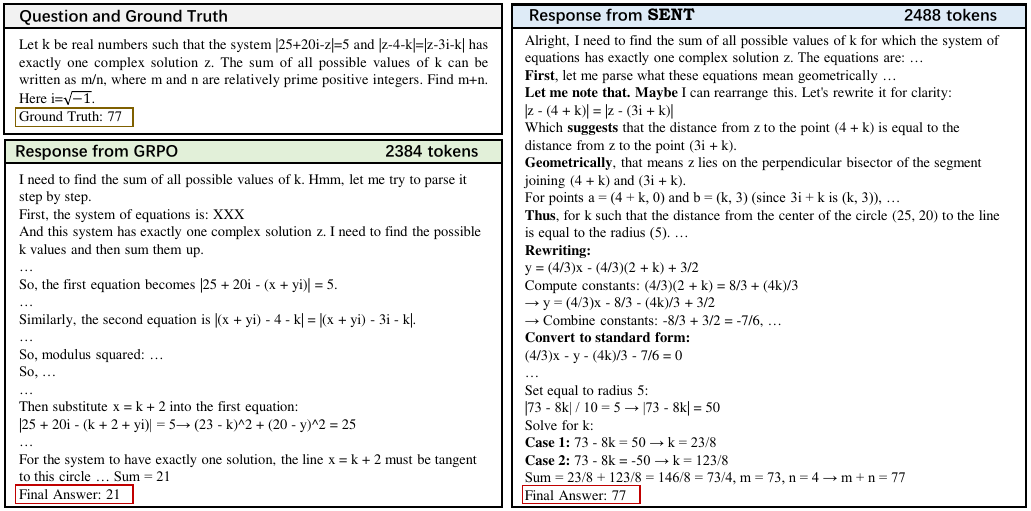}
\caption{The response comparison of specific prompt and response from GRPO and \method for one case. }
\label{fig:case}
\end{figure*}

\subsection{Case Study}

Fig.~\ref{fig:case} presents a qualitative comparison between responses generated by GRPO and \method. We observe that the GRPO-generated response is characterized by frequent sequential transition words such as "so," indicating a reliance on direct, linear logical progression toward the final answer. In contrast, the \method-generated response exhibits exploratory linguistic markers including hedging expressions (e.g., "maybe") and cognitive cues (e.g., "rewriting"), alongside explicit case-by-case analysis. \textbf{These linguistic patterns reflect a qualitatively different reasoning process—one characterized by hypothesis exploration, self-correction, and multi-path consideration rather than deterministic forward chaining.} This case study corroborates our quantitative findings, demonstrating that \method enhances reasoning capability by encouraging deeper, more exploratory cognitive processes that consider alternative solution paths, rather than prematurely committing to a single logical trajectory.

\section{Conclusion and Future Work}
\label{sec:conclusion}
In this work, we propose a semantic and token entropy-guided RL framework to enhance LLMs' reasoning. \method consists of two key components: First, we design a curriculum learning strategy based on semantic entropy to organize training data in a progressive difficulty order, facilitating more effective knowledge acquisition. Second, we introduce a fine-grained token-level entropy regularization objective that encourages exploration. By integrating curriculum learning at the semantic level with entropy regularization at the token level, our approach progressively enhances reasoning capabilities while effectively addressing the entropy collapse problem that limits reasoning improvement in conventional RLVR methods. Extensive experiments across $6$ mathematical benchmarks demonstrate the effectiveness of our method, achieving consistent improvements over baselines. Furthermore, cross-domain evaluation on LiveCodeBench validates the generalization, confirming that the reasoning enhancements transfer.

\textbf{Future Work:} Future directions involve enhancing training stability by incorporating empowerment-based optimization objectives, which provide principled mechanisms for controlled exploration and can mitigate the policy instability occasionally observed during training. Second, at the data organization level, we intend to explore causal reasoning frameworks to refine our curriculum design, moving beyond entropy-based difficulty estimation toward causally-informed data sequencing. These directions aim to further improve both the stability and scalability.


\bibliographystyle{IEEEtran}
\bibliography{tpami.bib}

\clearpage
\onecolumn

\appendix
\subsection{Broader Impact}
Our method offers substantial broader impact by addressing the fundamental challenge of entropy collapse in reinforcement learning for LLMs' reasoning. By strategically organizing training data through semantic entropy-guided curriculum learning and applying fine-grained token-level regularization, \method enables more efficient and stable policy optimization that maintains healthy exploration throughout training. This approach has promising implications for developing more capable reasoning systems across diverse domains, from mathematical problem-solving to complex decision-making tasks.

The dual-perspective optimization framework that combines data organization with algorithmic intervention provides a generalizable paradigm that could extend beyond reasoning tasks to other areas where exploration-exploitation balance is critical, such as code generation, planning, and multi-step problem solving. Our entropy-aware curriculum design principle offers insights for organizing training data in other machine learning contexts where task difficulty impacts learning dynamics.

Despite its strengths, our method has certain limitations. The effectiveness of semantic entropy-based curriculum learning depends on the quality of entropy estimation, which may vary across different types of reasoning tasks. Additionally, while we focus on token-level entropy characteristics, there is potential to explore more sophisticated granularities of analysis, such as reasoning step-level or sub-problem-level entropy patterns, which could lead to even more nuanced optimization strategies.

Future work will investigate extending our framework to other challenging domains such as multi-modal reasoning and long-context scenarios, where entropy dynamics may exhibit different characteristics. We also aim to explore adaptive curriculum strategies that can dynamically adjust difficulty progression based on real-time learning signals, further enhancing training efficiency and reasoning capabilities.

\subsection{Proofs}
\textbf{Proposition 1:} (Logit Change in Policy Gradient~\cite{NIPS2001_4b86abe4,cui2025entropy}). \textit{When updating policy via Policy Gradient with learning rate $\eta$, the logit difference satisfies:}
\begin{equation*}
\theta^{k+1}_{s_t,o_t} - \theta^k_{s_t,o_t} = \eta\cdot\pi^k_\theta(o_t|s_t)A_t
\label{eq:logit_difference}
\end{equation*}
\begin{proof}[Proof adapted from\cite{cui2025entropy}]
\begin{equation*}
\begin{aligned}
& \theta^{k+1}_{s_t,o_t} - \theta^k_{s_t,o_t} 
\\ &= \eta \cdot \nabla_{\theta_{s_t,o_t}} \mathcal{J}(\theta) \\
&= \eta \cdot \mathbb{E}_{o'_t \sim \pi^k_{\theta}(\cdot | s_t)} \left[ \nabla_{\theta_{s_t,o_t}} \log \pi^k_{\theta}(o'_t | s_t) \cdot A_t \right] \\
&= \eta \cdot \mathbb{E}_{o'_t \sim \pi^k_{\theta}(\cdot | s_t)} \left[ \frac{\partial \log \pi^k_{\theta}(o'_t | s_t)}{\partial \theta_{s_t,o_t}} \cdot A_t \right] \\
&= \eta \cdot \sum_{o'_t \in \mathcal{O}} \left[ \pi^k_{\theta}(o'_t | s_t) \cdot \left( \mathbbm{1}\{o_t = o'_t\} - \pi^k_{\theta}(o_t | s_t) \right) \cdot A_t \right] \\
&= \eta \cdot \pi^k_{\theta}(o_t | s_t) \cdot \Bigg[ \left(1 - \pi^k_{\theta}(o_t | s_t)\right) \cdot A_t - \sum_{o'_t \in \mathcal{O}, o'_t \neq o_t} \pi^k_{\theta}(o'_t | s_t) \cdot A_t \Bigg] \\
&= \eta \cdot \pi^k_{\theta}(o_t | s_t) \cdot \left[ A_t - \sum_{o'_t \in \mathcal{O}} \pi^k_{\theta}(o'_t | s_t) \cdot A_t \right] \\
&= \eta \cdot \pi^k_{\theta}(o_t | s_t) \cdot \left[ A_t - \mathbb{E}_{o'_t \sim \pi^k_{\theta}(\cdot | s_t)}[A_t] \right] \\
&= \eta \cdot \pi^k_{\theta}(o_t | s_t) \cdot \left[ A_t - \left( V^{\pi^k_{\theta}}(s_t) - V^{\pi^k_{\theta}}(s_t) \right) \right] \\
&= \eta \cdot \pi^k_{\theta}(o_t | s_t) \cdot \left[ A_t - 0 \right] \\
&= \eta \cdot \pi^k_{\theta}(o_t | s_t) \cdot A_t
\end{aligned}
\end{equation*}
\end{proof}

\textbf{Eq.~\ref{eq:L2}:}
\textit{When updating policy via our optimization objective $\mathcal{J}_{\text{SENT}}$ with learning rate $\eta$, the logit difference satisfies:}
\begin{equation*}
\hspace{-4mm}
\theta^{k+1}_{s_t,o_t} - \theta^k_{s_t,o_t}=
\eta\,\Big(\pi_\theta^k(o_t| s_t) A_t
-\beta_{\text{con}}\,
\nabla_\theta D_{\mathrm{KL}}(\pi_\theta^k\|\pi_{\mathrm{ref}})\Big)
\end{equation*}

\begin{proof}Following the same derivation as Proposition 1, the logit update satisfies:
\begin{equation*}
\begin{aligned}
& \theta^{k+1}_{s_t,o_t} - \theta^k_{s_t,o_t} 
\\ & = \eta \cdot \nabla_{\theta_{s_t,o_t}} \mathcal{J}_{\text{SENT}}(\theta) \\
&= \eta \cdot \nabla_{\theta_{s_t,o_t}} \Bigg[ \mathbb{E}_{o'_t \sim \pi^k_{\theta}(\cdot | s_t)} \left[ \log \pi^k_{\theta}(o'_t | s_t) \cdot A_t \right] - \beta_{\text{con}} D_{\text{KL}}(\pi^k_{\theta} \| \pi_{\text{ref}}) \Bigg] \\
&= \eta \cdot \mathbb{E}_{o'_t \sim \pi^k_{\theta}(\cdot | s_t)} \left[ \nabla_{\theta_{s_t,o_t}} \log \pi^k_{\theta}(o'_t | s_t) \cdot A_t \right]  - \eta \beta_{\text{con}} \nabla_{\theta_{s_t,o_t}} D_{\text{KL}}(\pi^k_{\theta} \| \pi_{\text{ref}}) \\
&= \eta \cdot \pi^k_{\theta}(o_t | s_t) A_t - \eta \beta_{\text{con}} \nabla_{\theta} D_{\text{KL}}(\pi^k_{\theta} \| \pi_{\text{ref}}) \quad \text{(Proposition. 1)} \\
&= \eta \left( \pi^k_{\theta}(o_t | s_t) A_t - \beta_{\text{con}} \nabla_{\theta} D_{\text{KL}}(\pi^k_{\theta} \| \pi_{\text{ref}}) \right)
\end{aligned}
\end{equation*}
\end{proof}
\subsection{Implementation Details of Baselines}
\label{sec:baseline-details}
We compare \method with 7 representative baselines.
including vanilla GRPO, directly adding entropy into the learning objective~\cite{haarnoja2018soft} (w/ En), adding an entropy-based advantage function~\cite{cheng2025reasoning} (w/ Adv), masking low-entropy tokens~\cite{wang2025beyond} (w/ Mask), clipping a small fraction of high-covariance tokens~\cite{cui2025entropy} (w/ Clip), constrain high covariance tokens~\cite{cui2025entropy} (w/ Cov) and adding a high-entropy reward for optimization (w/ High\_En).
\subsubsection{GRPO}
For the GRPO baseline, we follow the original GRPO objective defined in Eq.~\ref{eq:grpo}.

\subsubsection{w/ En}
w/ En is a method that directly adds an entropy regularization term to the learning objective. We implement this baseline by following the token-mean entropy regularization mechanism in VeRL. The resulting w/ En objective is:
\begin{equation*}
\label{eq:wen}
\mathcal{J}_{\mathrm{En}}(\theta)
= \mathcal{J}_{\mathrm{GRPO}}(\theta)
\;+\; \lambda \, \mathbb{E}\big[\mathcal{H}_t\big],
\end{equation*}

\subsubsection{w/ Adv}
For w/ Adv, we follow the original paper~\cite{cheng2025reasoning} to reproduce this entropy-based advantage shaping method. First, we compute token-level entropy $\mathcal{H}_t$ and construct a clipped shaping term:
\begin{equation*}
\psi(\mathcal{H}_t)=\min\!\left(\alpha \cdot 
\mathcal{H}_t^{\text{detach}},\, \frac{|A_t|}{\kappa}\right).
\end{equation*}

The shaped advantage is then defined as:
\begin{equation*}
A_t^{\text{shaped}} = A_t + \psi(\mathcal{H}_t).
\end{equation*}

The final w/ Adv objective is:
\begin{equation*}
    \begin{aligned}
\mathcal{J}_{\mathrm{Adv}}(\theta) 
&= \mathbb{E}_{(q,a)\sim \mathcal{D},\,\{o^i\}_{i=1}^G \sim \pi_{\theta_{\mathrm{old}}}(\cdot | q)} 
\\
& \Bigg[ \frac{1}{G} \sum_{i=1}^{G} \frac{1}{|o^i|} 
\sum_{t=1}^{|o^i|}
\Big( \min\!\big(r^i_t(\theta)\hat{A}^{i_{shaped}}_t,\, \mathrm{clip}(r^i_t(\theta),
& 1-\epsilon,\, 1+\epsilon)\hat{A}^{i_{shaped}}_t\big)
- \beta D_{\mathrm{KL}}(\pi_{\theta} \,\|\, \pi_{\mathrm{ref}}) \Big)
\Bigg].
    \end{aligned}
\end{equation*}

Following the original paper, we set the KL loss coefficient to 0, $\kappa$ to 2 and $\alpha$ to 0.4 in our implementation.
\subsubsection{w/ Mask}
For w/ Mask, we follow the original paper~\cite{wang2025beyond} by masking low entropy tokens during optimization. The objective is:

$$  \begin{aligned}&\mathcal{J}_{\mathrm{Mask}}^{\mathcal{B}}(\boldsymbol{\theta})=\mathbb{E}_{\mathcal{B}\sim\mathcal{D},(\boldsymbol{q},\boldsymbol{a})\sim\mathcal{B},\{\boldsymbol{o}^{i}\}_{i=1}^{\mathcal{G}}\sim\pi_{\boldsymbol{\theta}_{\mathrm{o l d}}}(\cdot|\boldsymbol{q})}\Biggl[\frac{1}{\sum_{i=1}^{G}\sum_{t=1}^{|\boldsymbol{o}^{i}|}\mathbb{I}\left[H_{t}^{i}\geq\tau_{\rho}^{\mathcal{B}}\right]}\sum_{i=1}^{G}\sum_{t=1}^{|\boldsymbol{o}^{i}|}\mathbb{I}\left[H_{t}^{i}\geq\tau_{\rho}^{\mathcal{B}}\right]\\&\cdot\min\left(r_{t}^{i}(\boldsymbol{\theta})\hat{A}_{t}^{i},\operatorname{c l i p}\left(r_{t}^{i}(\boldsymbol{\theta}),1-\epsilon,1+\epsilon\right)\hat{A}_{t}^{i}\right)\Biggr], s.t.0<\left|\left\{\boldsymbol{o}^{i}\mid\mathrm{i s\_{e} q u i v a l e n t}(\boldsymbol{a},\boldsymbol{o}^{i})\right\}\right|<G,\\ \end{aligned}  $$
where $\mathcal{B}$ denotes a  micro-batch sampled from the training dataset $\mathcal{D}$, and $\tau_{\rho}^{\mathcal{B}}$ is the threshold selecting the top-$\rho$ high-entropy tokens. In our implementation, we set $\rho=0.2$, following the original paper.
\subsubsection{w/ Clip}
w/ Clip strategy clips a small fraction of high-covariance tokens from policy gradient updates. In practice, We first compute covariances defined in Eq.~\ref{eq:covariance}, then randomly select $r \cdot N$ tokens whose covariance falls in $[\omega_{\mathrm{low}}, \omega_{\mathrm{high}}]$:

\begin{equation*}
I_{\mathrm{clip}} \sim \mathrm{Uniform}\Big(i \mid \mathrm{Cov}(y_i) \in [\omega_{\mathrm{low}}, \omega_{\mathrm{high}}], \lfloor r \cdot N \rfloor \Big),
\end{equation*}
where $I_{\mathrm{clip}}$ denotes the indices of clipped tokens, $r$ is the clip ratio, and $\omega_{\mathrm{low}}, \omega_{\mathrm{high}}$ are high covariance bounds. We follow the original paper and set the clip ratio $r = 2\times10^{-4}$ with covariance bounds $\omega_{\mathrm{low}} = 1$ and $\omega_{\mathrm{high}} = 5$.

The final policy loss is:
\begin{equation*}
L_{\mathtt{Cov}}(\theta) =
\begin{cases}
\mathbb{E}_t \big[ \frac{\pi_\theta(y_t \mid \mathbf{y}_{<t})}{\pi_{\theta_\mathrm{old}}(y_t \mid \mathbf{y}_{<t})} A_t \big], & t \notin I_{\mathrm{clip}}, \\
0, & t \in I_{\mathrm{clip}}.
\end{cases}
\end{equation*}
\subsubsection{w/ Cov}
This strategy applies a KL penalty to high-covariance tokens. We first compute covariances defined in Eq.~\ref{eq:covariance}, then select the top-$k$ proportion of tokens:

\begin{equation*}
I_{\mathrm{KL}} = \{ i \mid \mathrm{Rank}(\mathrm{Cov}(y_i)) \le k \cdot N \},
\end{equation*}
where $k \ll 1$ is the proportion of tokens to penalize. Following the original paper, we set $k$ to 0.0002. The final policy loss is:
\begin{equation*}
L_{\mathtt{Cov}}(\theta) =
\begin{cases}
\mathbb{E}_t \big[ \frac{\pi_\theta(y_t \mid \mathbf{y}_{<t})}{\pi_{\theta_\mathrm{old}}(y_t \mid \mathbf{y}_{<t})} A_t \big], & t \notin I_{\mathrm{KL}}, \\
\mathbb{E}_t \Big[ \frac{\pi_\theta(y_t \mid \mathbf{y}_{<t})}{\pi_{\theta_\mathrm{old}}(y_t \mid \mathbf{y}_{<t})} A_t 
- \beta \, \mathbb{D}_{\mathrm{KL}}(\pi_{\theta_\mathrm{old}}(y_t \mid \mathbf{y}_{<t}) \,\|\, \pi_\theta(y_t \mid \mathbf{y}_{<t})) \Big], & t \in I_{\mathrm{KL}}.
\end{cases}
\end{equation*}
\subsubsection{w/ High\_En}
For w/ High\_En, we add a high entropy reward to the GRPO objective. Specifically, the objective is:
\begin{equation*}
\mathcal{J}_{\mathrm{High\_En}}(\theta) = \mathcal{J}_{\mathrm{GRPO}}(\theta) + \lambda \sum_{i,t} \mathbb{I}[H_t^i \ge \tau] H_t^i,
\end{equation*}

\subsection{Details on Experimental Design and Results}
\subsubsection{Experimental Setup}
To clearly illustrate the experimental setup and ensure reproducibility, we provide a detailed description of the training hyperparameters and evaluation hyperparameters for our experiments in Table~\ref{tab:training_params} and Table~\ref{tab:eval_params}. Parameters not explicitly listed in the table follow the default configurations of the VeRL~\cite{sheng2025hybridflow} framework. Additional special hyperparameters specific to certain baselines are configured strictly according to their original papers, as mentioned in Section~\ref{sec:baseline-details}.

As shown in Table~\ref{tab:training_params}, We conduct RL training on three models of different sizes and capabilities including DeepSeek-R1-Distill-Qwen-1.5B, Qwen2.5-Math-7B, and Qwen3-14B. For training data, we use DAPO-MATH-17K datasets~\cite{yu2025dapo}, an elaborately curated math dataset. We reproduce all baseline methods and apply consistent hyperparameters across different approaches within the VeRL platform. For training efficiency, we set the number of training epochs to 5 for the 1.5B model and 3 for both the 7B and 14B models. Specially, we set the coefficient of KL divergence loss to 0.001 for GRPO and 1.0 for w/ Cov and \method, while setting it to 0 for methods that do not use KL regularization. Similarly, we set the entropy coefficient to 0.001 for methods that employ entropy regularization, such as w/ En and w/ High En, and 0 for the remaining approaches. For evaluation, as detailed in Table~\ref{tab:eval_params}, we employ consistent inference configurations across all benchmarks to ensure fair comparisons. 
Codes are available at: \codesite.
\begin{table}[t]
    \caption{Training hyperparameters used in our experiments.}
    \renewcommand{\arraystretch}{1.2}
    \setlength{\tabcolsep}{5pt}
    \centering
    \begin{tabular}{lccc}
        \toprule
        \textbf{Hyperparameters} 
        & \textbf{DeepSeek-R1-Distill-Qwen-1.5B} 
        & \textbf{Qwen2.5-Math-7B} 
        & \textbf{Qwen3-14B} \\
        \midrule
        \textbf{epochs}         
        & 5           
        & 3   
        & 3 \\
        
        \textbf{grpo rollout size}    
        & 8         
        & 8      
        & 8 \\
        
        \textbf{temperature}    
        & 1.0          
        & 1.0      
        & 1.0 \\

        \textbf{top-p}          
        & 1.0           
        & 1.0         
        & 1.0 \\

        \textbf{top-k}
        & -1           
        & -1         
        & -1 \\

        \textbf{learning rate}  
        & 1e-6      
        & 1e-6      
        & 1e-6 \\
        
        \textbf{max response length}
        & 2048           
        & 2048         
        & 2048 \\
        
        \textbf{global batch size} 
        & 256         
        & 256         
        & 256 \\
        
        \textbf{ppo mini-batch size} 
        & 128         
        & 128         
        & 128 \\ \hline

        \textbf{kl loss coefficient} \\
        \quad \textbf{GRPO}
        & 0.001         
        & 0.001         
        & 0.001 \\
        \quad \textbf{w/ Cov and w/ \method}
        & 1.0         
        & 1.0         
        & 1.0 \\
        \quad \textbf{w/o kl loss} 
        & 0
        & 0
        & 0 \\ \hline

        \textbf{entropy coefficient} \\ 
        \quad \textbf{w/ entropy loss} 
        & 0.001
        & 0.001
        & 0.001 \\
        \quad \textbf{w/o entropy loss} 
        & 0
        & 0
        & 0 \\
        \bottomrule
    \end{tabular}
    \label{tab:training_params}
\end{table}

\begin{table}[t]
    \caption{Evaluation hyperparameters used in our experiments.}
    \renewcommand{\arraystretch}{1.2}
    \setlength{\tabcolsep}{5pt}
    \centering
    \begin{tabular}{lccc}
        \toprule
        \textbf{Hyperparameters} 
        & \textbf{Value} \\
        \midrule       
        \textbf{temperature}          
        & 1.0 \\

        \textbf{top-p}                 
        & 1.0 \\

        \textbf{max response length}        
        & 8000 \\

        \textbf{k}        
        & 1, 8, 16, 32 \\
        \bottomrule
    \end{tabular}
    \label{tab:eval_params}
\end{table}

\subsubsection{Benchmarks}
In our experiments, we conduct extensive validation on following six challenging mathematical reasoning benchmarks to comprehensively evaluate model performance.

\begin{itemize}
    \item \textbf{AIME 2024 \& 2025}~\cite{codeforcesamerican}: A collection of 30 problems from the American Invitational Mathematics Examination 2024/2025, a prestigious high school mathematics competition featuring challenging multi-step problems across various mathematical domains.
    
    \item \textbf{AMC 2023}~\cite{amc23}: A set of 40 problems from the 2023 American Mathematics Competitions.
    
    \item \textbf{MATH500}~\cite{hendrycks2measuring}: A 500-problem subset from the MATH dataset covering seven subjects including Algebra, Geometry, Number Theory, and Precalculus at competition mathematics level.
    
    \item \textbf{OlympiadBench}~\cite{he2024olympiadbench}: A challenging benchmark that provides Olympiad-level, bilingual, multimodal scientific problems designed to evaluate advanced mathematical and scientific reasoning in large language models. In our experiments, We only use the OE\_TO\_maths\_en\_COMP subset consisting of 674 open-ended, text-only English mathematics competition problems.
    
    \item \textbf{Minerva}~\cite{lewkowycz2022solving}: Minerva is a benchmark designed to evaluate the mathematical and quantitative reasoning capabilities of LLMs. It consists of 272 problems sourced primarily from MIT OpenCourseWare courses, covering advanced STEM subjects such as solid-state chemistry, astronomy, differential equations, and special relativity at the university and graduate level.
\end{itemize}

\subsubsection{Additional Results of Entropy Changes} 
\begin{figure}[t]
\centering
\includegraphics[width=0.4\linewidth]{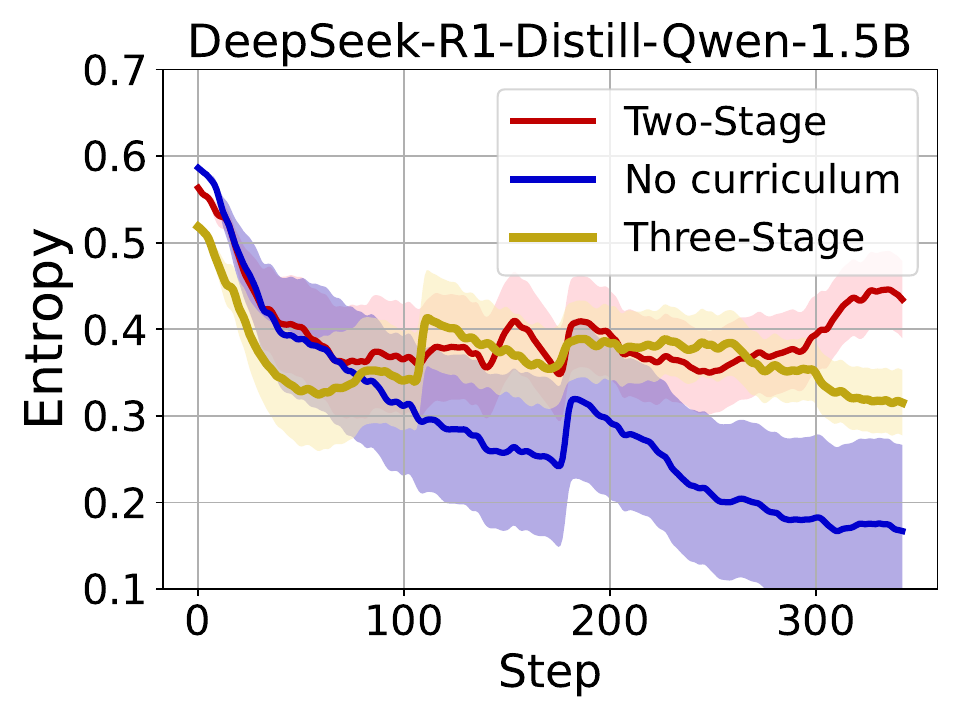}
\includegraphics[width=0.4\linewidth]{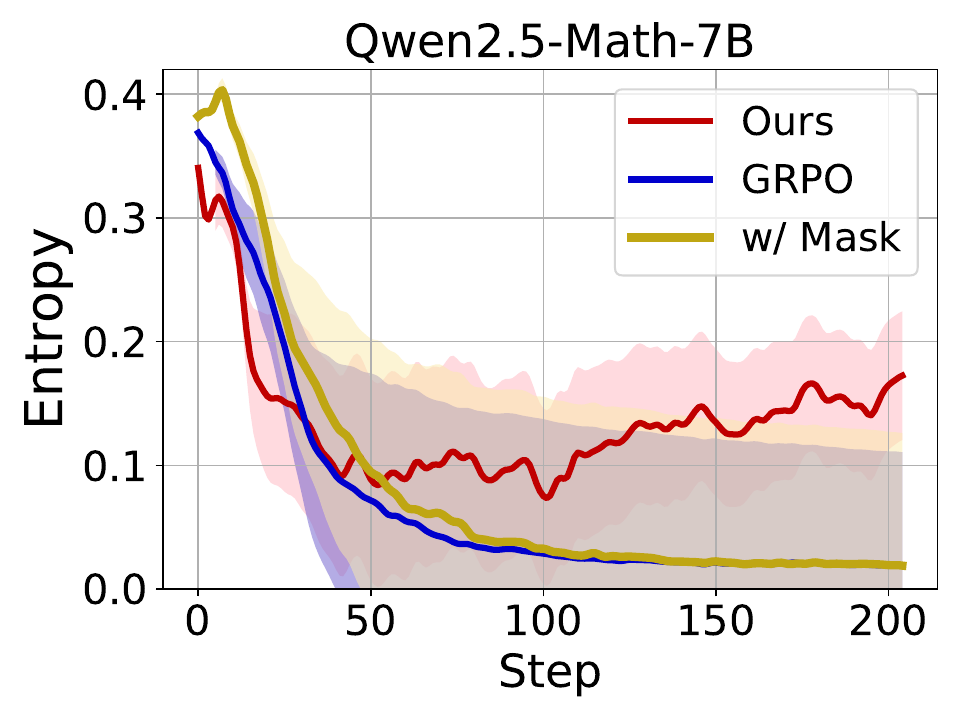}
\caption{The learning curves of entropy changes during learning process from $1.5$B and $7$B models. w/ Mask means GRPO with mask low entropy tokens. The shadow of line is the standard error.}
\label{fig:entropy_appendix}
\end{figure}

To further analyze entropy changes during the post-training process, we compare different curriculum designs and baseline methods. As illustrated in Fig.~\ref{fig:entropy_appendix}, the results reveal distinct patterns across configurations. Among curriculum designs, the no-curriculum approach leads to precipitous entropy drops, while both two-stage and three-stage curricula successfully maintain entropy stability throughout training. Notably, the two-stage curriculum consistently preserves higher entropy levels, thereby sustaining stronger exploration capacity. In the 7B model experiments, our method stands as the only approach that completely avoids entropy collapse, whereas both GRPO and the w/ Mask exhibit entropy collapse during training.

These results directly validate our research objective: effectively mitigating entropy collapse to prevent the dramatic reduction in policy exploration that limits reasoning capabilities. The consistent maintenance of healthy entropy levels across different model scales and benchmark tasks demonstrates that our dual-perspective optimization framework—combining semantic curriculum learning with fine-grained token-level regularization—successfully addresses this fundamental challenge in RLVR-based reasoning enhancement.

\subsubsection{Performance in $14$B Base Model}
\begin{figure}[t]
\centering
\includegraphics[width=0.32\linewidth]{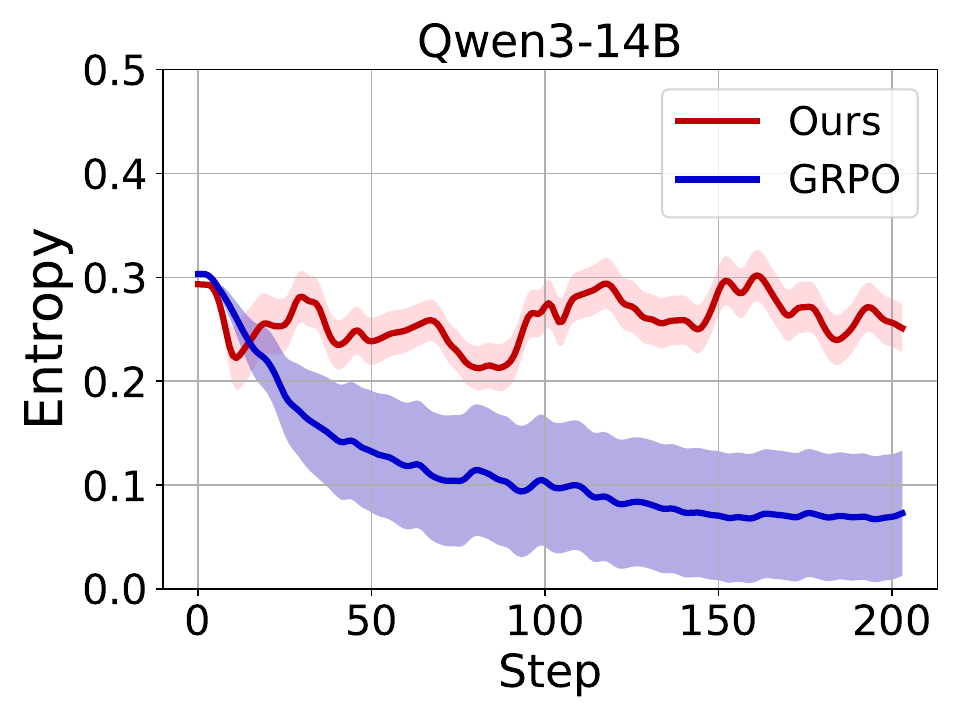}
\includegraphics[width=0.32\linewidth]{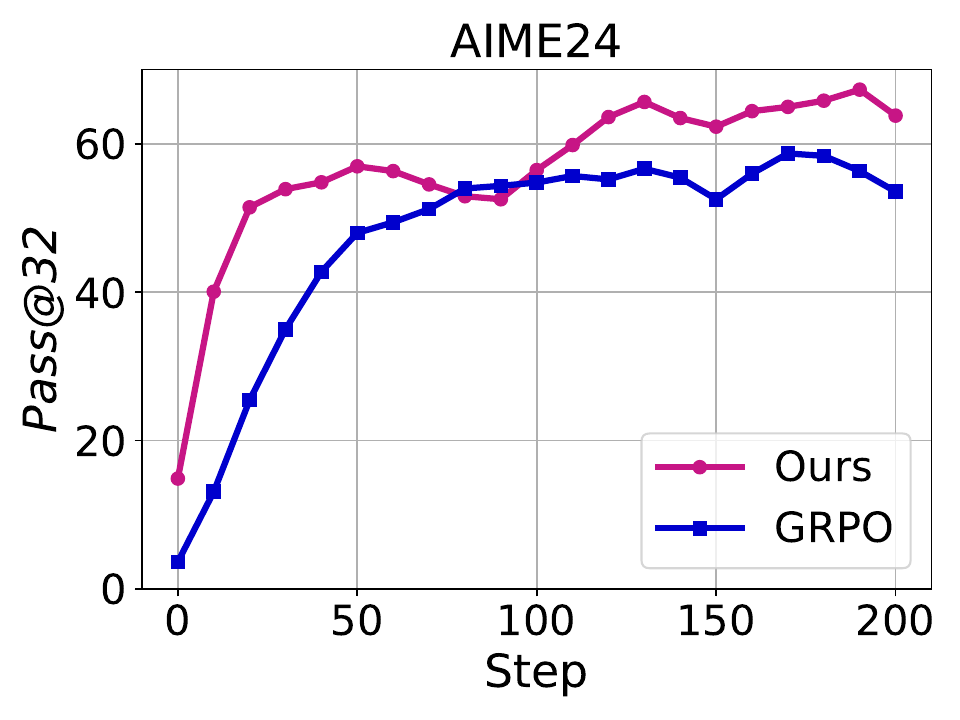}
\includegraphics[width=0.32\linewidth]{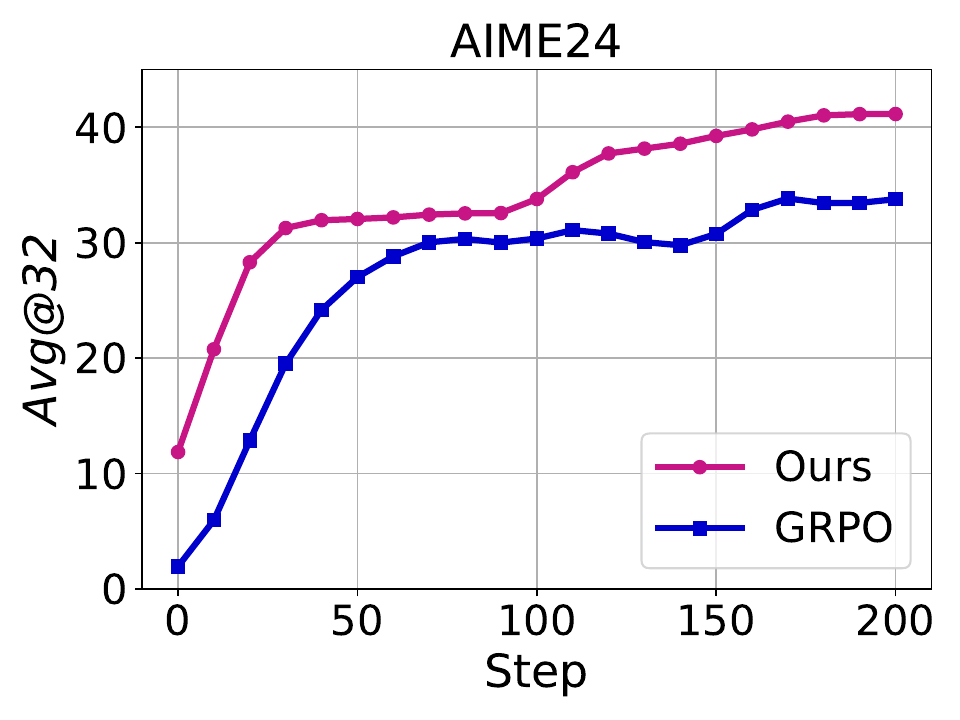}
\caption{The learning curves of entropy changes, pass@16 and avg@16 during learning process $14$B models.The shadow of line is the standard error.}
\label{fig:14b_appendix}
\end{figure}

We analyze the performance of \method compared with GRPO on Qwen3-14B base model, as shown in Fig.~\ref{fig:14b_appendix}. The entropy dynamics reveal that our method maintains stable entropy throughout training, avoiding both the collapse observed in GRPO and potential entropy explosion. This stable entropy progression indicates healthy exploration maintenance at larger model scales. Furthermore, monitoring performance on the AIME24 validation set during training shows that \method achieves superior progressive learning compared to GRPO, with more consistent and steady improvement across training iterations. These results demonstrate that our optimization framework effectively scales to larger models, enabling reasoning enhancement.

\subsubsection{Case Study} 
\begin{figure*}
\centering
\includegraphics[width=1\linewidth]{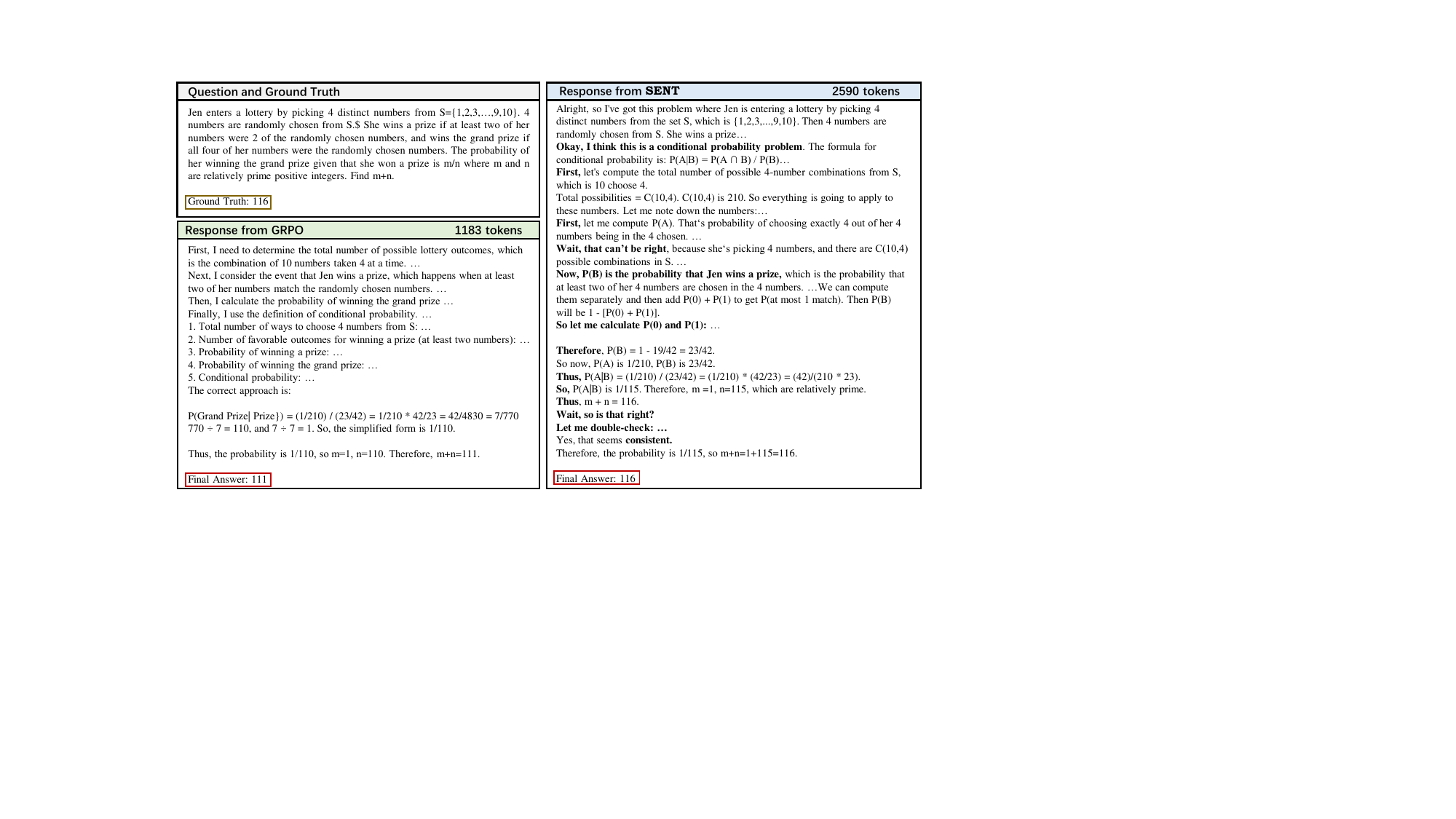}
\caption{The response comparison of specific prompt and response from GRPO and \method. }
\label{fig:case_2}
\end{figure*}

We provide an additional response comparison in Fig.~\ref{fig:case_2}. The GRPO response is notably shorter and follows a straightforward logical reasoning process without verification steps, ultimately leading to an incorrect answer. In contrast, \method generates a more comprehensive response that incorporates reflective reasoning throughout the solution process. Notably, the model performs self-verification and double-checking of intermediate steps, a critical capability for robust reasoning. These observations further validate the effectiveness of our proposed framework in improving LLMs' reasoning capabilities.

\subsubsection{Comparison with OpenPangu} To further validate the competitive performance of our method, we conducted a comparison with the OpenPangu-embedded-1B-v1.1 model~\cite{chen2025pangu}. As detailed in Table \ref{tab:openpangu}, SENT significantly outperforms OpenPangu across all six benchmarks under the Avg@16 metric. On average, SENT surpasses OpenPangu by $8.88$ points, confirming that our entropy-guided curriculum learning strategy effectively unlocks the reasoning potential of small-scale models.
\begin{table}[h]
    \centering
    \caption{Performance comparison of Avg@16 between OpenPangu-embedded-1B-v1.1 and SENT (Ours) based on DeepSeek-R1-Distill-Qwen-1.5B. We bold the best scores.}
    \label{tab:openpangu}
    \renewcommand{\arraystretch}{1.2} 
    \setlength{\tabcolsep}{8pt}       
    \begin{tabular}{l c c}
        \toprule
        \textbf{Benchmark} & \textbf{OpenPangu-1B} & \textbf{w/ \method (Ours)} \\
        \midrule
        AIME24        & 17.92 & \textbf{25.21} \\
        AIME25        & 18.96 & \textbf{21.46} \\
        AMC23         & 46.25 & \textbf{70.16} \\
        MATH500       & 76.78 & \textbf{81.74} \\
        OlympiadBench & 36.30 & \textbf{42.99} \\
        Minerva       & 14.57 & \textbf{21.25} \\
        \midrule
        \textbf{Avg.} & 35.13 & \textbf{44.01} \\
        \bottomrule
    \end{tabular}
\end{table}

\subsection{Experimental Platforms and Licenses}
\subsubsection{Platforms}
All experiments of this approach are implemented on two Intel Xeon Platinum 8480+ CPUs and eight NVIDIA H800 GPUs.

\subsubsection{Licenses}
In our code, we have utilized the following libraries, each covered by its respective license agreements:
\begin{itemize}
    \item VeRL (Apache License 2.0)
    \item RAY (Apache License 2.0)
    \item TensorDict  (MIT License)
    \item Flash‑Attn (BSD 3-Clause "New" or "Revised" License)
    \item vLLM (Apache License 2.0)
\end{itemize}

\end{document}